\documentclass[11pt]{article}

\usepackage[preprint]{acl}
\newcommand{\balanced}{\textsc{Nested}}
\newcommand{\matched}{\textsc{Equal-Count}}
\newcommand{\heuristic}{\textsc{First-Symbol}}
\newcommand{\open}{{\fboxsep=0pt\fcolorbox{lightgray}{white}{\textcolor{purple}{\textbf{(}}}}}
\newcommand{\close}{{\fboxsep=0pt\fcolorbox{lightgray}{white}{\textcolor{teal}{\textbf{)}}}}}
\newcommand{\EOS}{\texttt{EOS}}
\newcommand{\BOS}{\texttt{BOS}}
\usepackage{multirow} 
\usepackage{enumitem}
\usepackage{hyphenat} 
\usepackage{graphicx}
\usepackage{subcaption}
\usepackage{booktabs}
\usepackage{amsmath}
\usepackage{enumitem}
\usepackage{times}
\usepackage{latexsym}
\usepackage{amssymb}

\usepackage{booktabs, multirow, makecell, xcolor, pifont}
\usepackage[normalem]{ulem}
\newcommand{\cmark}{\textcolor{green!55!black}{\ding{51}}}
\newcommand{\xmark}{\textcolor{red!70!black}{\ding{55}}}
\newcommand{\auxM}[1]{\textcolor{purple}{\textbf{#1}}}
\newcommand{\auxE}[1]{\textcolor{teal}{\textbf{#1}}}

\usepackage[]{todonotes}
\makeatletter
\newcommand*\iftodonotes{\if@todonotes@disabled\expandafter\@secondoftwo\else\expandafter\@firstoftwo\fi}
\makeatother

\usepackage[compact]{titlesec}

\usepackage[T1]{fontenc}

\usepackage[utf8]{inputenc}

\usepackage{microtype}

\usepackage{inconsolata}

\usepackage{graphicx}

\title{Can Interpretation Predict Behavior on Unseen Data?}

\author{
  \textbf{Victoria R. Li\textsuperscript{1}},
  \textbf{Jenny Kaufmann\textsuperscript{1}},
  \textbf{Tian Qin\textsuperscript{1}},\\
  \textbf{Martin Wattenberg\textsuperscript{1}},
\textbf{David Alvarez-Melis\textsuperscript{1}},
  \textbf{Naomi Saphra\textsuperscript{1,2}}
\\
\\
  \textsuperscript{1}Harvard University,
  \textsuperscript{2}Boston University
\\
  \small{
    \textbf{Correspondence:} \href{mailto:vrli@alumni.harvard.edu}{vrli@alumni.harvard.edu}, \href{mailto:nsaphra@bu.edu}{nsaphra@bu.edu}
  }
}

\begin{document}
\maketitle

\begin{abstract}
   Interpretability research often  predicts model responses to targeted mechanistic \textit{interventions}. But can we predict responses to unseen \textit{input data}? We propose and demonstrate this alternate objective by using model internals to predict their out-of-distribution (OOD) behavior. We train hundreds of Transformers on simple synthetic tasks, where perfect in-distribution accuracy is compatible with multiple OOD generalization rules. We successfully use attention patterns---observed only on in-distribution data---to predict which rule each model follows on OOD data. Our experiments decouple the mechanistic faithfulness of our interpretation from its predictive value; ablations reveal such internal patterns can \textit{suppress} rather than \textit{support} the rule they predict, showing observational analysis can forecast behavior even when causal analysis fails to support a simple cause-effect link. Our findings are a proof-of-concept for a new interpretability objective: understanding model internals to predict behavior and assess reliability under distribution shift.
\end{abstract}

\section{Introduction}\label{sec:intro}

When do we understand a system? One standard, that of the classic scientific method~\citep{sep-scientific-method}, requires \textit{testable predictions of behavior under unseen conditions}.  Interpretability research is often evaluated by predicting the effect of a test-time mechanistic intervention~\citep{geiger2024causalabstractiontheoreticalfoundation} such as activation steering~\citep{subramani2022extractinglatentsteeringvectors,tan2025analyzinggeneralizationreliabilitysteering,liu2024incontextvectorsmakingcontext} or patching~\citep{makelov2023subspacelookingforinterpretability,kramár2024atpefficientscalablemethod,todd2024functionvectorslargelanguage,vig2020}. In contrast, interpretability researchers rarely infer a model's rules by predicting its outputs on unseen \textit{data} inputs. We will demonstrate the latter objective.

This work proposes a new mission for interpretability. We argue that progress in model understanding should be measured not only by causal-mechanistic intervention, but also by predicting model behavior on unseen data.

If we could predict model behavior under distribution shift, we would unlock new interpretability applications. Well-understood instruments come with engineering tolerances---specified edge cases where the instrument may fail. By providing tolerances for AI models, we might guide training, offer recommendations for reliable use, or flag potential deployment failures. For example, structures associated with specific languages and tasks may provide clues for whether a model can  compose them to fluently handle a given task in a particular language. Can we use interpretation to predict \mbox{a model's response to unseen data?}

\begin{figure}
    \centering
    \includegraphics[width=\linewidth]{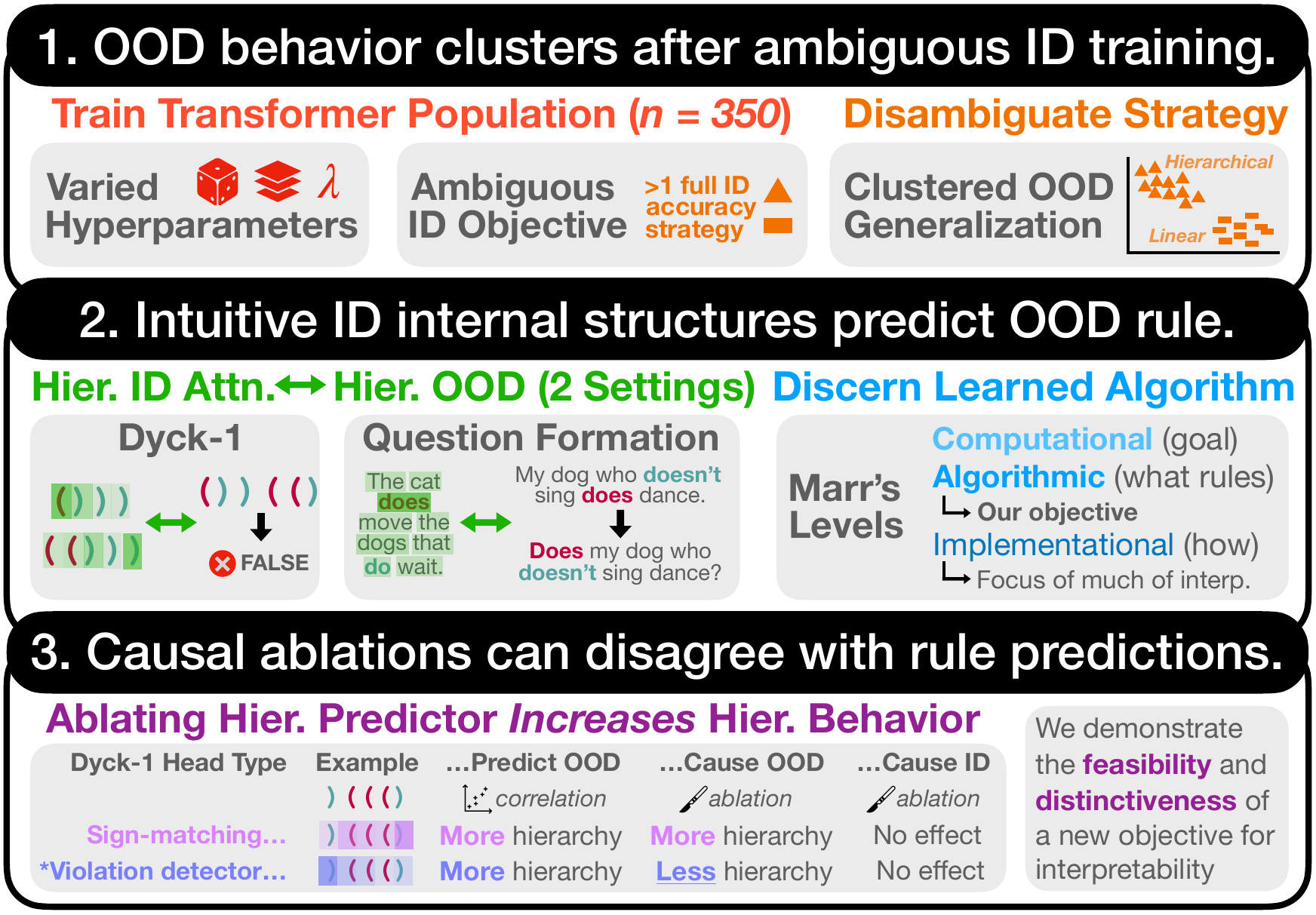}
    \caption{\textbf{Our findings.}}
    \label{fig:placeholder}
    \vspace{-1.5em}
\end{figure}

Using synthetic tasks, we show that simple analysis of attention patterns can reveal a model's generalization rule \textit{even if} they are not used straightforwardly in its implementation. In other words, our interpretations may not be \textit{faithful} in a causal, mechanistic sense---but they nonetheless reveal informative traces left by the algorithm. Interpretability-based intuitions then allow us to guess the algorithm being executed, and therefore to predict how models treat unseen data. 

We view the model as a whole and use its internals to understand resulting outputs---a \textit{holistic interpretability} approach. We wish to predict a model's \textit{out-of-distribution}~(OOD) behavior using only its internal representations of \textit{in-distribution} (ID) data.
To this end, we train a large set of Transformer-based classifiers with perfect ID validation accuracy and diverse OOD behavior in a synthetic setting.
Our model population is trained on data with an ambiguous classification rule: models can achieve perfect ID accuracy using either a parenthesis \textbf{counting} rule (\matched{}) or a hierarchical parenthesis \textbf{nesting} rule (\balanced{}).
We use an OOD test set to determine which rule each model follows. (See Appendix \ref{apx:glossary} for a glossary.)

In Marr's terms \citep{marr2010vision}, our objective targets the \textit{computational} and \textit{algorithmic}---rather than implementational---levels of analysis. We treat internals not as causal mechanisms but as signatures hinting at what structures the model represents for its rule. Full model understanding requires multiple levels of analysis.

Overall, motivated by epistemic concerns and actionable interpretability objectives, we train a population of 270 models to classify Dyck-1 parentheses strings (Sec.~\ref{sec:experiments}), with an additional autoregressive English question-formation setting (Appendix \ref{apx:qf}). We find (Figure \ref{fig:placeholder}):
\begin{itemize}[leftmargin=*, topsep=0pt, parsep=0pt, itemsep=2pt, partopsep=0pt]
    \item \textbf{Independently trained models cluster around systematic generalization rules.} We visualize model OOD judgments, finding clusters that implement linear and hierarchical strategies to varying degrees (Sec.~\ref{sec:generalization-behavior}). In the Dyck-1 setting, some unregularized training runs learn a  heuristic instead of a valid rule, but only apply this heuristic OOD (Sec.~\ref{sec:rule-identification}). Models can pass through transient heuristic phases in training, suggesting that vestigial circuits still affect OOD judgments. In addition to weight decay, model depth and seed influence rule learning (Sec.~\ref{sec:twinstudiesfactors}).
    \item \textbf{OOD generalization rules are intuitively predicted by internal representations of ID data.} We identify heads that encode hierarchical structure in their attention activations (Sec.~\ref{sec:head-types}). In the Dyck-1 setting, we show that models with these heads usually apply \balanced{} on unseen OOD data (Sec.~\ref{sec:predicting-ood}). In the second, autoregressive question-formation setting, we also find a hierarchical internal attention pattern that correspondingly predicts hierarchical OOD generalization (Sec.~\ref{sec:qf-main}).
    \item \textbf{Internal structures predict generalization rules even when the rule's implementation doesn't rely on them.} We differentiate circuits \emph{causally} necessary in implementing a rule from those \textit{correlated} with the rule across a model population. 
    In our autoregressive setting (Appendix \ref{apx:qf}), ablation tests leave OOD behavior unchanged. 
    In our Dyck-1 setting, 
    we find that our interpretation's predictive power is clearly decoupled from its causal faithfulness. \balanced{} is associated with multiple types of hierarchical attention heads,  but some types suppress---rather than support---the \balanced{} rule (Sec.~\ref{sec:ablation}). Furthermore, the effect of ablation is only weakly correlated between ID and OOD data conditions, calling into question the robustness of causal interpretability findings (Sec.~\ref{sec:datadependent}). 
\end{itemize}

\subsection{Philosophical Motivation}
\label{sec:philosophy}

The interpretability of attention is contentious \citep{bibal2022attention}: specific attention patterns are rarely necessary for model performance, either in theory \citep{dyckcasestudy} or in practice \citep{jain-wallace-2019-attention,serrano-smith-2019-attention}. At its heart, this debate is philosophical---about what makes an interpretation faithful. We take a different epistemic position: our objective is to predict a model's response to unseen inputs, not to mechanistic interventions.

\paragraph{Causal interpretability and its limits.} In modern interpretability, causality is often held as the gold standard for evaluating understanding. Researchers causally test each module by ablating it and measuring damage to model performance. However, the resulting findings are often compatible with multiple analyses which can only be differentiated by expensive, precise ablations. Moreover, proposed interpretations may not transfer to new settings~\citep{stander_grokking_2024, zhangnanda, makelov2023subspacelookingforinterpretability}. By focusing on causal intervention alone to test the faithfulness of explanations, the literature on understanding attention has produced skepticism and fierce debate~\citep{bibal2022attention,dyckcasestudy}.

\paragraph{Correlational analyses in the sciences.} While interpretability often demands causal tests, other sciences frequently use correlational analyses. In genetics, for example, correlational studies can link inherited traits like red hair and painkiller tolerance \citep{liem2005increased}. While causal studies may seem epistemically sounder in principle, engineered approaches like single-gene editing can sabotage unrelated processes through complex interactions.\footnote{\citet{haapaniemi2018crispr} provide a famous example of how targeted gene edits can produce spurious results. They detail how CRISPR gene editing can activate p53, a tumor suppressor. As a result, experimentalists using the method inadvertently select for cells with defective p53 responses, forming a confounder for research on cancer genetics.} Why should it be any easier to isolate mechanisms in highly nonlinear neural networks?

\paragraph{Related work on causal predictions OOD.} Some existing work uses causal analysis to predict model computations on OOD inputs. The visual illusion constructed by \citet{gurnee2026modelsmanipulatemanifoldsgeometry} validated a causal attribution by proposing a specific failure mode based on inspecting attention patterns. \citet{huang2025internalcausalmechanismsrobustly} showed that ID causal features predict errors better than ID non-causal features when used to simulate OOD scenarios. By contrast, our setting demonstrates that some causal experiments can mislead as to the predictive power of representational structures.

\paragraph{Interpretability and scientific realism.} Of the many controversies in philosophy of science, few are more central than the rivalry between realism and instrumentalism \citep{okasha2016philosophy}. Realists claim that concepts used in scientific models, from quarks to gravity, are specific objects and forces acting on the world \citep{putnam1975mathematics}. Instrumentalists, by contrast, argue that science's epistemic goal is not to uncover fundamental truths about the world, but to make predictions about their observable outcomes \citep{duhem1954aim}. In the philosophy of mind, instrumentalists may even deny that internal beliefs, desires, or intentions are real phenomena, while still acknowledging that these concepts might help predict a person's behavior \citep{churchland1981eliminative,dennett1991real,dennett1989intentional}. Our objective is an instrumentalist one: rather than insisting on a true hierarchical mechanism, we treat hierarchical structure as part of our scientific model of network behavior, sufficient if we can infer it from the traces it leaves.

Like other observational studies, we leverage correlations over model populations to test our understanding. Hidden representations can \textit{proxy} a model's algorithm even when they are not employed in its implementation, enabling our computational---rather than implementational---level of analysis \citep{marr2010vision}.

\section{Methods and Experiment Setup}
\label{sec:experiments}

To predict models' OOD behavior from their representation of ID data, we first create models with varied OOD generalization to study. We train all these models on a dataset compatible with at least two distinct OOD generalization rules (Sec.~\ref{sec:data-preliminaries}). We study the variation produced by training many models with different hyperparameters and random seeds (Sec.~\ref{sec:attention}).

\subsection{Data setting}
\label{sec:data-preliminaries}

Our setting is inspired by work on systematic generalization from ambiguous training rules~\citep{mccoy-etal-2020-syntax,mccoy2020bertsfeathergeneralizetogether,structuralgrokking}. Our primary dataset is based on the Dyck-1 parentheses-balancing task~\citep{suzgun2019,ebrahimi-etal-2020-self,structuralgrokking,dyckcasestudy}. Unlike standard parentheses-balancing settings, our training dataset is compatible with either \matched{}, an unordered counting rule, or \balanced{}, a hierarchical  parentheses-balancing rule. 

Our Dyck-1 models are classifiers, not sequence generators.
They verify that the input follows some rule and output a binary class $y \in \{\texttt{True},\texttt{False}\}$. Each input is an $n$-length sequence $s = s_1s_2\ldots s_n$, where $s_i \in \{  \open{} ,  \close{} \}$. Using $\mathbf{1}(\cdot)$ as the indicator function, let the number of open \open{} and close \close{} tokens that appear from index $1$ to $j$  be $o(j) = \sum_{i=1}^j \mathbf{1}(s_i =  \open )$ and $c(j) =  \sum_{i=1}^j \mathbf{1}(s_i =  \close )$, respectively. Each rule labels $s$ as follows.
\begin{itemize}[leftmargin=*, topsep=0pt, parsep=0pt, itemsep=2pt, partopsep=0pt]
    \item \matched{} is \texttt{True} if $s$ has the same number of open and close parentheses:
    \begin{equation}\label{eq:matched} o(n) = c(n)
    \end{equation}

    \item \balanced{} is \texttt{True} if $s$ is a recursively nested tree. All \balanced{} sequences are also \matched{}, but in addition to Eq.~\ref{eq:matched}, \balanced{} $s$ fulfills:
    \begin{equation}\label{eq:nested}
     \forall j \in \{1, \ldots, n\} \hspace{5mm} o(j) \geq c(j)
    \end{equation} 

\end{itemize} 
Our training set is compatible with both \matched{} and \balanced{}: each input satisfies \textit{both} or \textit{neither} of Equations \ref{eq:matched} and \ref{eq:nested}. Thus, \textit{a model can perfectly classify ID data by either rule}.  We test which rule each model learned using an OOD set of sequences that fulfill Equation \ref{eq:matched} but not Equation \ref{eq:nested}: they have the same number of \open{} and \close{} tokens, but not in a properly nested order.  We define accuracy according to \balanced{}, so a model has 100\% OOD accuracy if it labels all OOD examples \texttt{False}.

\paragraph{Implementation}
We create ID sequences where \balanced{} and \matched{} agree and OOD sequences where they disagree. Each sequence is generated randomly with length sampled $n \sim \textrm{Binomial}(40,0.5)$ (see Appx.~\ref{apx:dataset}). The 1K OOD examples fulfill \matched{} but not \balanced{} (e.g.,  \close\close\open\open\open\close \,).  
Our 1M-example train set, and 1K-example ID validation set are label-balanced, containing $50\%$ negative examples  which fulfill neither \matched{} nor \balanced{} (e.g., \open \open \close \close \close \,) and $50\%$ positive examples which fulfill both \matched{} and \balanced{} (e.g.,  \open\close\open\open\close\close \,).

\subsection{Models and attention}\label{sec:attention}

We train Transformers with causal self-attention and input length $L = 42$. Each sequence $s$ consists of a \BOS{} (beginning-of-sequence) token at $s_0$, a sequence of $n \leq 40$ parentheses, an \EOS{} (end-of-sequence) token at $s_{n+1}$, and $L-n-2$ padding tokens after \EOS{}. We output a binary class $\hat{y} \in \{\texttt{True},\texttt{False}\}$ at the final layer \EOS{} token.

Because our models output a classification label at the \EOS{} token, we consider attention activations at the \EOS{} index $n+1$.  For a given head's attention activations $A \in \mathbb{R}^{L \times L}$ on input $s$, we define $a_{\EOS{}}(i)$ as attention to token $s_i$ at index $i \in \{1, \ldots, n\}$:
\begin{equation}
    a_{\EOS{}}(i) = A_{n+1, i}.
\end{equation}

\paragraph{Implementation}

We train a population of classifier models based on the minGPT architecture~\cite{minGPT} with hidden dimension 64. Following \citet{vaswani2023attentionneed}, we use reshaping for multi-head attention, so the model's overall parameter count is the same regardless of per-layer head count $W$.  We set the learning rate $\eta = 0.0001$ with no dropout. All trained models stabilize to an ID validation accuracy of at least 99\% after at most 900K training examples (Appx. Fig.~\ref{fig:training-convergence}).

\begin{figure*}[t]
\centering
    \begin{subfigure}[b]{0.49\linewidth}
        \centering
        \includegraphics[height=4.6cm]{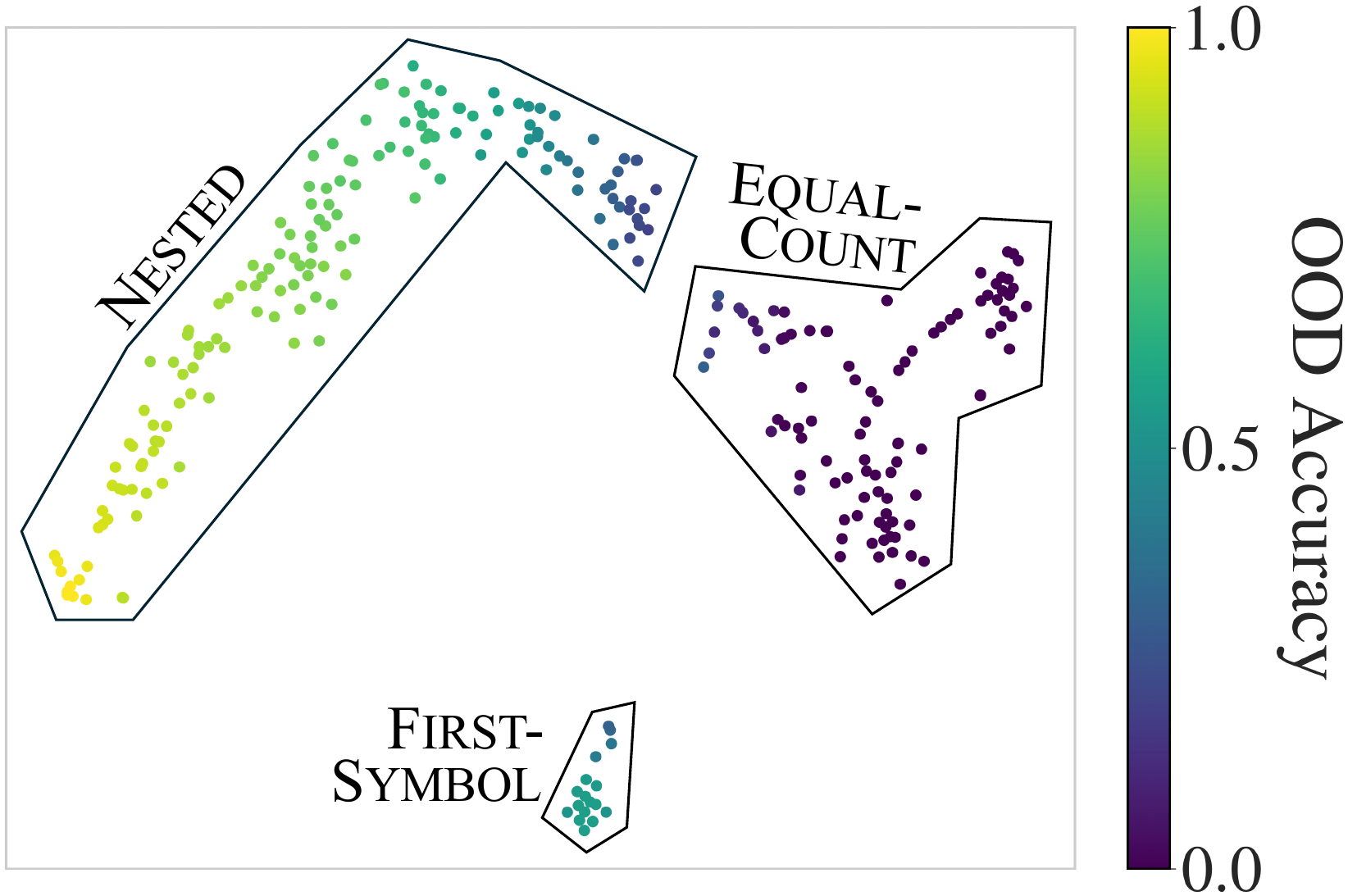}
        \caption{Dyck-1 classification.}\label{fig:tsne-dyck}
    \end{subfigure}
    \hfill
    \begin{subfigure}[b]{0.49\linewidth}
        \centering
        \includegraphics[height=4.6cm]{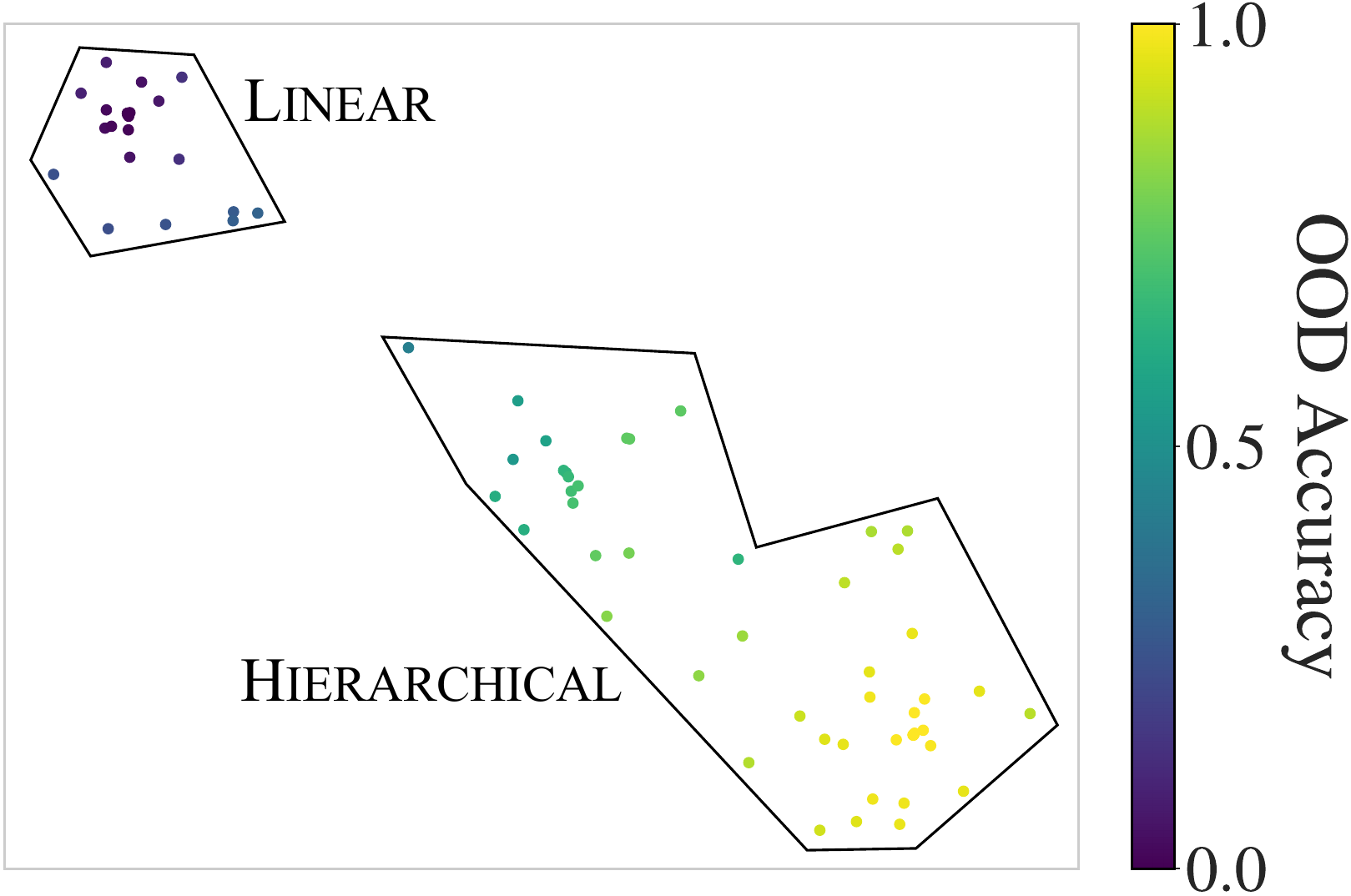}
        \caption{Question formation.}\label{fig:tsne-qf}
    \end{subfigure}
    \caption{\textbf{Independently trained models cluster around systematic generalization rules.} t-SNE of each model's OOD output probabilities. In Dyck-1 (a), models separate into \balanced{}, \matched{}, and a small \heuristic{} heuristic cluster. In question formation (b), models separate into a \textsc{Linear} cluster (low OOD accuracy) and a \textsc{Hierarchical} cluster (high OOD accuracy).}
    \vspace{-1em}
    \label{fig:tsne-ood-prob}
\end{figure*}

We grid sweep over other hyperparameters to create a diverse model population. We train models with depths $D \in \{1, 2, 3\}$ layers and widths $W \in \{2, 4\}$ attention heads per layer. We also vary optimizer weight decay $\lambda \in \{0, 0.001, 0.01\}$. Per combination, we train models with 5 random seeds for weight initialization and 3 for dataset shuffle order. This sweep results in $15$ models per hyperparameter configuration and 270 models in total.

\subsection{Additional Question Formation Setting}\label{sec:qf-setup} To complement our classifier experiments, we also study autoregressive language models on English question formation~\cite{mccoy-etal-2020-syntax,linzen-etal-2016-assessing,qin2024itreedatadrives}. Models read a declarative sentence and generate the corresponding yes/no question, which requires moving the \emph{matrix} (root-clause) auxiliary to the front. ID training data is constructed so that the linear ``move the first auxiliary'' rule and the hierarchical ``move the matrix auxiliary'' rule produce identical outputs; OOD examples distinguish these strategies (Appendix Table~\ref{tab:strategies}). We analyze a population of 79 autoregressive Transformers, trained to perfect ID accuracy. Appendix~\ref{apx:qf} includes more detail.

\section{Generalization Behavior}
\label{sec:generalization-behavior}

All Transformer models achieve high ID validation accuracy, but their OOD behaviors vary.

\subsection{Models cluster by systematic rule}\label{sec:rule-identification}

We visualize all models according to their OOD output probabilities (Fig.~\ref{fig:tsne-ood-prob}). In Dyck-1 (Fig.~\ref{fig:tsne-dyck}), one cluster has near-0\% OOD accuracy and another has high OOD accuracy, supporting existing claims~\citep{qin2024itreedatadrives,zhao2024distscaling} that \textit{rule-following is a categorical phenomenon, not a continuum}. Outside these two expected \matched{} and \balanced{} clusters, a third cluster of near-identical judgments emerges---a heuristic-based cluster. In question formation (Fig.~\ref{fig:tsne-qf}), models likewise separate into a linear-rule cluster (low OOD accuracy) and a hierarchical-rule cluster (high OOD accuracy), supporting the categorical nature of rule-following.

\subsubsection{The heuristic cluster}\label{sec:heuristic}

In the Dyck setting, the outlier cluster represents a simple heuristic, \heuristic{}: a sequence is \texttt{True} if its first symbol is \open{} and \texttt{False} if its first symbol is \close{}. This heuristic correctly labels all positive examples, but only half of negative examples (Appx. Table~\ref{tab:firstsymbol}). Although some models use this heuristic OOD, they do \textit{not} apply it to ID data or they would not achieve high ID validation accuracy.

\begin{figure}[!ht]
    \centering
    \includegraphics[width=0.7\linewidth]{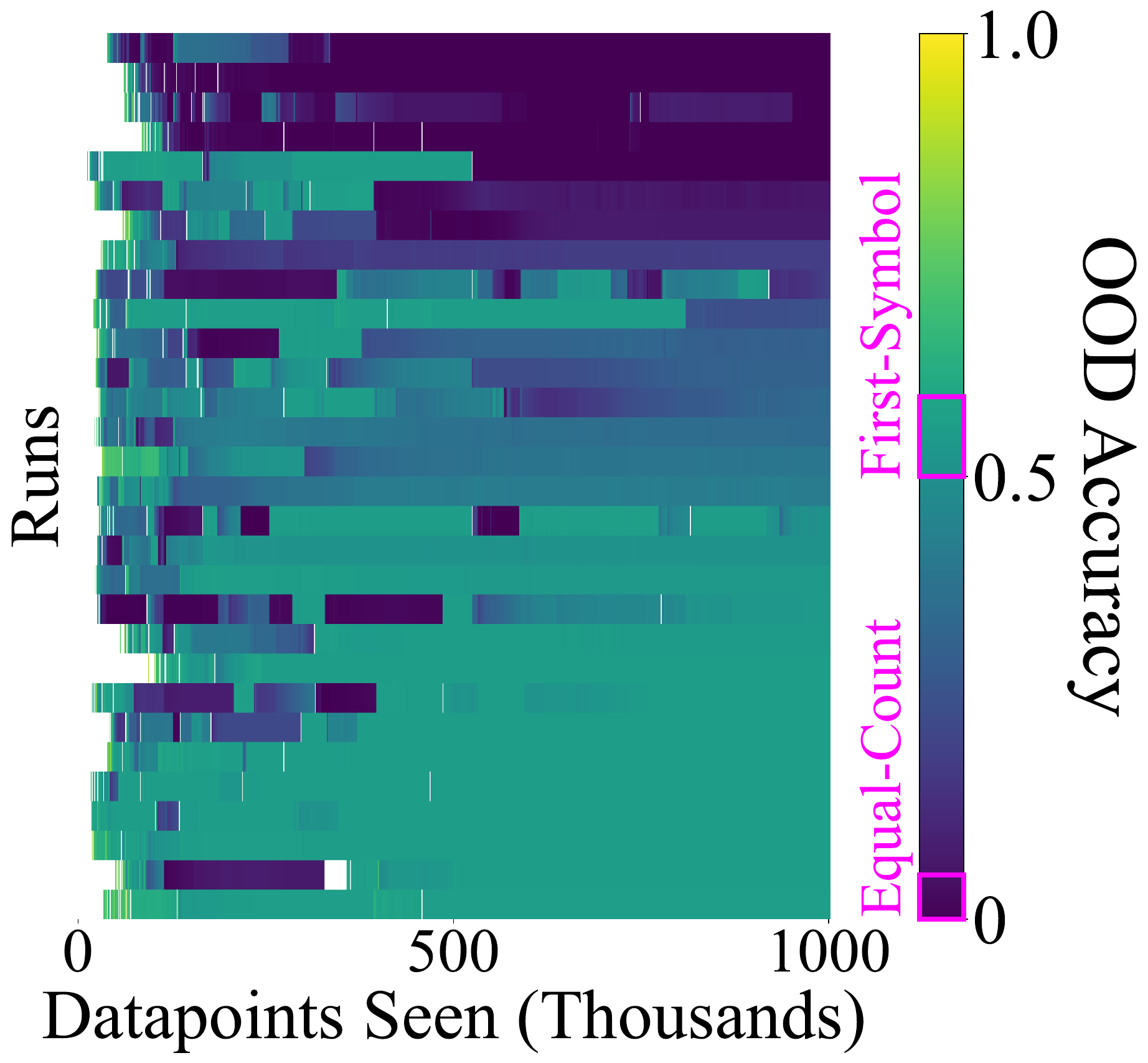}
    \caption{\textbf{Unregularized models experience transient heuristic training stages.} Intermediate checkpoints of unregularized $1$-layer models, sorted by final OOD accuracy. 
    Checkpoints that cannot model the training distribution (ID accuracy $< 0.99$) are shown in white cells. The color bar is marked at \heuristic{} and \matched{} accuracy levels. (More analysis in Appx.~\ref{sec:trdy}.)}
    \label{fig:heuristic-phase}
    \vspace{-1.2em}
\end{figure}

We hypothesize that the \heuristic{} heuristic is implemented by a vestigial circuit, which the model has learned to suppress on ID---but not OOD---data. Vestigial circuits are mechanisms needed early but not later in training. Prior literature~\citep{Tessier_2022,10247950,doshi2024grokgrokdisentanglinggeneralization} suggested that vestigial circuits are pruned by weight decay, which aligns with our observations of the \heuristic{} cluster: Note the \heuristic{} circuit only governs $1$-layer models trained \textit{without} weight decay, whereas all $1$-layer regularized models follow \matched{}. As further evidence of vestigiality, 1-layer \matched{} models (Fig.~\ref{fig:heuristic-phase}) often pass through apparent heuristic phases while training, during which OOD accuracy matches that of \heuristic{} models. These findings suggest that \emph{a vestigial circuit---with no detectable impact on ID behavior---can still affect OOD judgments.}

\subsection{Factors in rule selection}\label{sec:twinstudiesfactors}

Varying hyperparameters for the Dyck setting, we find the dominant factors in model rule selection are model depth and weight decay. Appx.~\ref{sec:factors} details other factors, including an extension of existing findings~\citep{abnar2020transferringinductivebiasesknowledge,mccoy-etal-2020-syntax,tran2018importancerecurrentmodelinghierarchical, saphra-lopez-2020-lstms} that recurrent architectures favor hierarchical structure.

\paragraph{Model depth} Appx. Fig.~\ref{fig:wd} shows 2- and 3-layer models can learn either \balanced{} or \matched{},  depending on hyperparameters and random seed.
In contrast, 1-layer models only learn \heuristic{} or \matched{}. Shallow models do not learn hierarchy, while 2-layer models tend more toward hierarchy than 3-layer models. This inverted U of hierarchical inductive bias by depth mirrors previous work~\cite{structuralgrokking}.

\paragraph{Weight decay} Regularization pushes models toward more consistent systematic rules, concentrating OOD behavioral distributions (Appx. Fig.~\ref{fig:wd}). In our setting, non-zero weight decay increases preference for \matched{} in 1- and 3-layer models but for \balanced{} in 2-layer models. These findings imply regularization promotes systematic rules over memorization, but \textit{which} rule is simplest depends on hyperparameters, complicating reductionist narratives of universally simple rules.

\section{Interpretability Predicts Behavior}

We inspect the attention activations of each head and use intuitive explanations of how ID inputs are processed to predict model decisions on unseen OOD inputs. Models following \balanced{} display hierarchical attention patterns on ID data; despite shared outputs, they can employ subtly different internal mechanisms with different causal roles. Hierarchical attention patterns predict that a model will follow \balanced{} \emph{even when} these patterns do not directly implement it. Appx.~\ref{apx:breakdown} confirms that attention patterns add predictive value beyond hyperparameter settings alone.

\subsection{Hierarchical attention patterns}
\label{sec:head-types}

Certain heads exhibit interpretable attention patterns at the \EOS{} token. Specifically, these patterns indicate whether a token position $j$ follows the condition in Equation \ref{eq:nested} that $o(j) \geq c(j)$. We will connect these intrinsically hierarchical heads to  the \balanced{} generalization rule.

\subsubsection{Tracking violations} 
\label{sec:tracking_depth}

We identify attention patterns that encode an input's hierarchical structure. These patterns attend to a token at position $j$ depending on if it violates Equation \ref{eq:nested}, i.e., if $o(j) < c(j)$. If a sequence contains any such \textbf{violation token}, the sequence is not \balanced{}. In particular, \textit{all}  examples in the OOD set---\matched{}, but not \balanced{}---must have at least one violation token. Intuitively, attention patterns which track these violations can signal that a model is closer to \balanced{}.

\begin{figure*}[!t]
    \centering
\begin{subfigure}{0.46\linewidth}
\includegraphics[width=\linewidth]{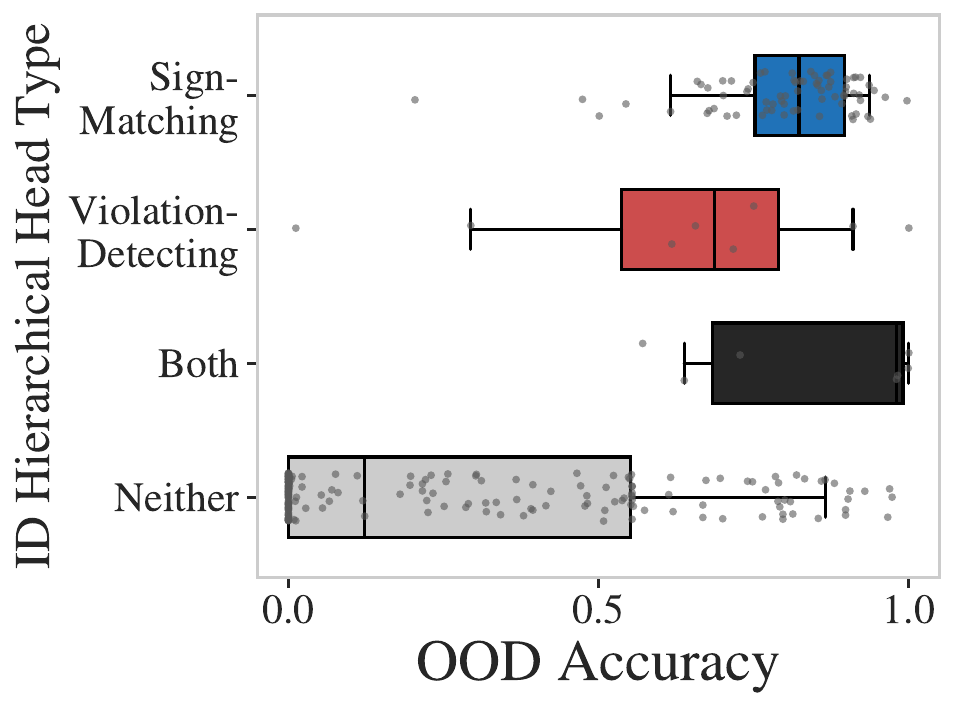}
\caption{Dyck-1: OOD accuracy of 2- and 3-layer models split by ID hierarchical head type. \emph{Two} subtypes of hierarchical head---sign-matching and violation-detecting---predict higher OOD accuracy than models with neither.}\label{fig:boxplot_dyck_split}
\end{subfigure}
\hfill
\begin{subfigure}{0.46\linewidth}
\includegraphics[width=\linewidth]{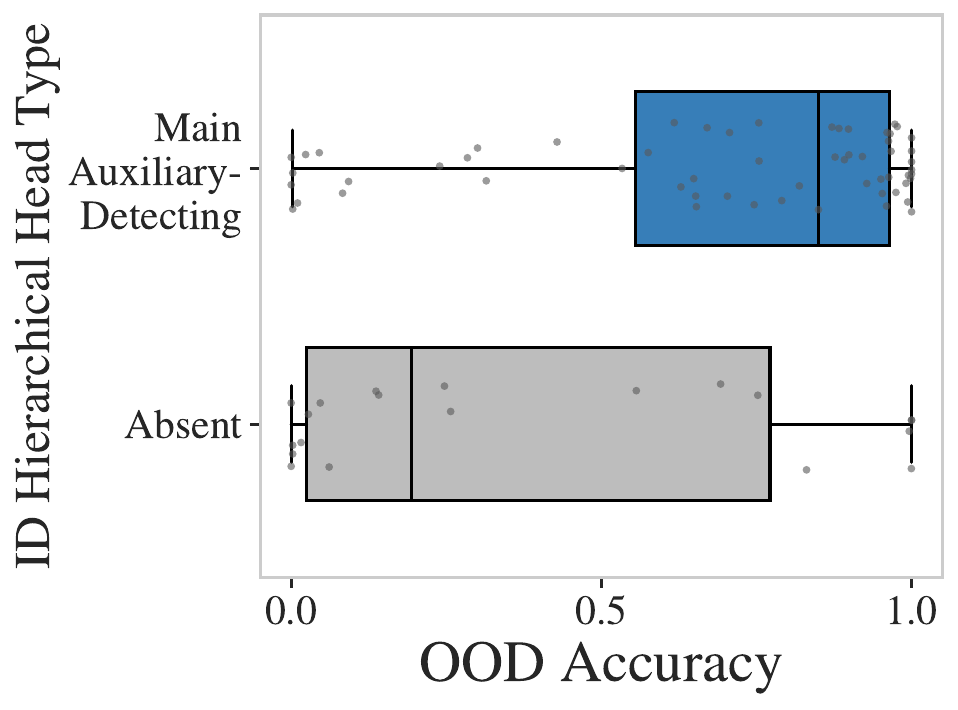}
\caption{Question formation: OOD accuracy split by presence of an ID \emph{Main Auxiliary-Detecting} head, which attends from the question-formation readout position to the matrix auxiliary. See more details in Appx.~\ref{apx:qf}.}\label{fig:boxplot_qf_split}
\end{subfigure}
\caption{\textbf{Hierarchical attention patterns predict hierarchical OOD generalization in two settings.} ID hierarchical heads, of either subtype, correlate with hierarchical OOD behavior in both our settings.}\label{fig:headtype_predicts_ood}
\end{figure*}

There are two ways for an attention output to \textbf{track violations} on a given sequence: a head can preferentially attend either to violation or to non-violation tokens. For input $s$,
\begin{itemize}[leftmargin=*, topsep=0pt, parsep=0pt, itemsep=2pt, partopsep=0pt]
\item An attention head favors \emph{violation tokens} on $s$ if there exists threshold $t > 0$ such that:
\begin{equation}
    \forall j \in \{ 1, \dots, n \}: a_{\EOS{}}(j) \geq t \text{ iff } o(j) < c(j)
\label{eq:negative-depth}\end{equation}

 \item An attention head favors \emph{non-violation tokens} on $s$ otherwise, i.e., there exists $t > 0$ such that:

 \begin{equation}
     \forall j \in \{ 1, \ldots, n \}: a_{\EOS{}}(j) \geq t \text{ iff } o(j) \geq c(j)
 \label{eq:nonneg-depth}\end{equation}
\end{itemize}
Examples of each pattern are shown in Appendix Fig.~\ref{fig:attn_examples_appx}. We say an attention head is \textbf{tracking violations on a given input} if it favors \textit{either} violation \textit{or} non-violation tokens on that input.

\subsubsection{Hierarchical head types} 

Some heads reliably track Equation \ref{eq:nested} violations on each input, making them hierarchical heads across a dataset. We define a head as a \textbf{hierarchical head} on a given dataset if it tracks violations (Appendix Figure \ref{fig:attn_examples_appx}) on at least 80\% of relevant sequences---those containing both violation and non-violation tokens.
Heads which track violations on ID examples typically behave as hierarchical heads OOD (i.e., 77\% of the time). We find two hierarchical head types:
\begin{itemize}[topsep=0pt, parsep=0pt, itemsep=2pt, partopsep=0pt]
    \item \textbf{Violation-detecting heads} consistently  assign more attention to violations of Equation \ref{eq:nested}, i.e., positions preceded by more \close{} than \open{} tokens.
    \item \textbf{Sign-matching heads} behave differently depending on whether the final token follows Eq. \ref{eq:nested}. If the final token violates Eq. \ref{eq:nested}, the head will attend to violations; if the final token follows Eq. \ref{eq:nested} (as in all OOD sequences), the head will attend to non-violations.
\end{itemize}

Among models with ID hierarchical heads, 81\% have sign-matching heads, 9\% have violation-detecting heads, and 8\% have both; only 2\% of models with hierarchical heads have neither subtype. We therefore focus on these subtypes, which cover almost all ID hierarchical heads.
 
\subsection{Hierarchical attention on ID data \textit{predicts} hierarchical rules on OOD data}

\label{sec:predicting-ood}

We return to our overarching question: Can these model internals intuitively suggest which function the model implements, thereby predicting its treatment of the unseen OOD distribution? If we claim to understand a model, we should be able to predict its behavior under unseen conditions. We now demonstrate that our attention-based intuitions can, in fact, reliably predict OOD model behavior.

As seen in Fig.~\ref{fig:boxplot_dyck_split}, Dyck models which contain at least one ID hierarchical head---of either type---are more likely to follow the \balanced{} rule. This result holds separately across 2- and 3-layer models (Appx.~\ref{apx:breakdown}). These findings confirm our intuitive hypothesis: ID hierarchical representational structure is associated with OOD hierarchical generalization. These ID internal patterns, along with other ID attention statistics, can predict the OOD rule of held-out models: the hierarchical head is one instance of a broader ID predictive signal (Appx.~\ref{apx:classifier}).

We find that 1-layer models, which notably \emph{do not} learn the \balanced{} rule (Appx. Fig.~\ref{fig:wd}), possess no hierarchical heads. In fact, hierarchical heads do not occur in the first layer of any model---possibly explaining why our models only learn \balanced{} if they have multiple layers.

All OOD sequences end with $o(n) = c(n)$, so a \textit{sign-matching} head attends to non-violation tokens. We do not observe any \textit{ID violation-detecting heads} switch to favoring non-violation tokens on OOD sequences. However, 25\% of \textit{ID sign-matching heads} do switch to consistently favoring violations OOD. Given that a mechanism can behave so differently OOD, it would be challenging to describe any internal mechanism in a way that applies across new domains. Nonetheless, these structures provide a \textit{holistic} understanding of the algorithm implemented by the model. This inferred algorithm, unlike mechanistic interpretations tied to particular internal patterns, applies across distribution shift.

 \begin{figure*}[!ht]
\begin{center}
\includegraphics[width=0.8\textwidth]{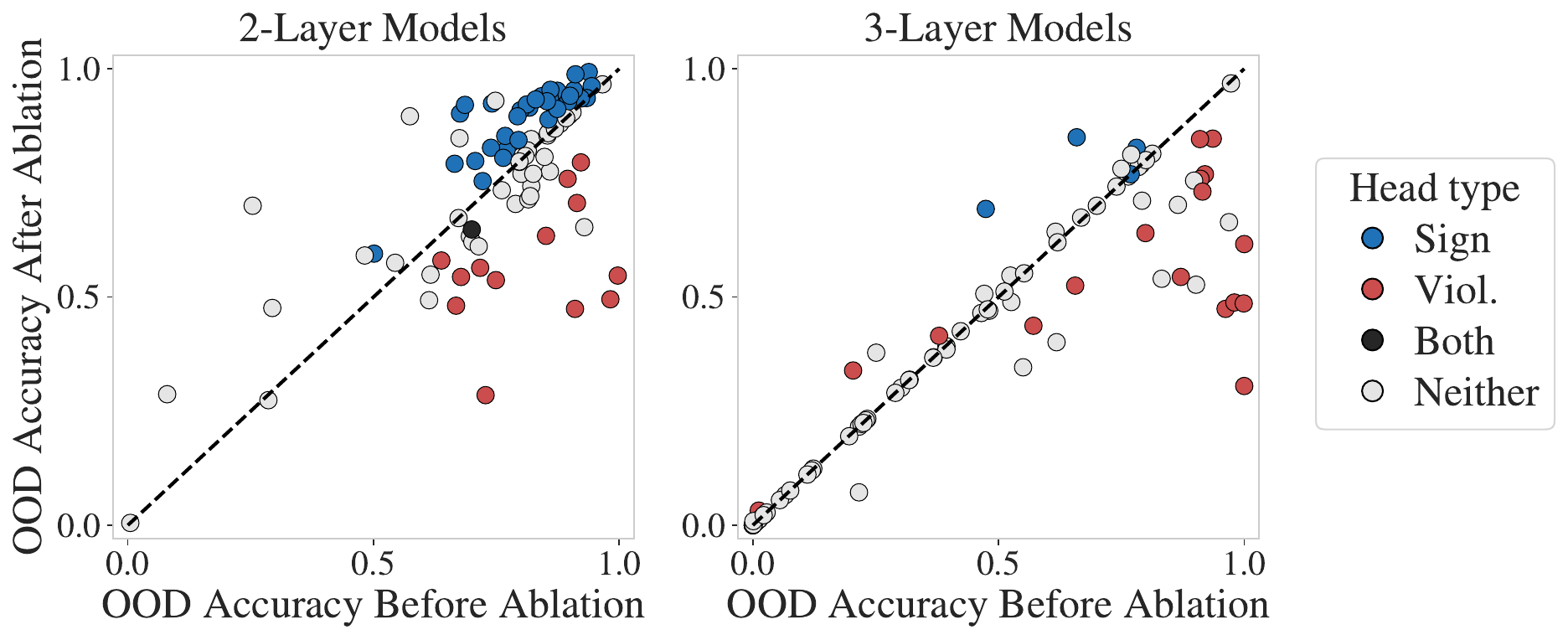}
\vspace{-3mm}
\end{center}
\caption{\textbf{Some hierarchical attention patterns damage the implementation of hierarchical rules (Dyck-1).} OOD accuracy before and after applying uniform attention ablation. Each point is a single model, colored by presence of an OOD sign-matching and/or violation-detecting head. Ablation damages OOD performance below the diagonal but improves it above. Although both head types are correlated with \balanced{}, only violation detection causally supports the rule; sign-matching attention \emph{suppresses} the rule.}\label{fig:scatter_plot_multilayers}
\vspace{-3mm}
\end{figure*}

\subsubsection{Question formation setting}\label{sec:qf-main}

In the question-formation population we also find an ID attention pattern that predicts hierarchy.
\begin{itemize}[topsep=0pt, parsep=0pt, itemsep=2pt, partopsep=0pt] \item \textbf{Main Auxiliary-Detecting heads} concentrate attention on the matrix (root-clause) auxiliary rather than the embedded-clause one.
\end{itemize}

Concretely, an ID sentence in this setting is ``\textit{My unicorn does move the dogs that do wait}''. Here, the ID output is ``Does my unicorn move the dogs that do wait?'' A Main Auxiliary-Detecting head for this sentence places substantial attention---more than $0.2$ of its attention mass---on the matrix auxiliary ``does'' (the hierarchically-correct target, as opposed to the embedded-clause auxiliary ``do'') just before the model outputs the question. We particularly identify heads in the first three layers of each model that exhibit this behavior on at least 80\% of ID examples. The presence of such an ID hierarchical head in a question-formation model strongly corresponds to OOD behavior with hierarchical question-formation output (Fig.~\ref{fig:boxplot_qf_split}). Like in the Dyck setting, we can make held-out predictions of OOD generalization based on ID attention-associated signals (Appx.~\ref{apx:classifier}). 

\subsection{Hierarchical attention may not \textit{cause} hierarchical rules on OOD data} 
\label{sec:ablation}

In mechanistic interpretability, faithfulness is  often evaluated through causal intervention.
In the Dyck setting, hierarchical heads correlate with the \balanced{} rule, but correlation alone doesn't establish them as causal mechanisms. To determine causality, we intervene on attention activations and examine how model performance responds. These experiments demonstrate that an attention pattern might correlate with a systematic rule without supporting it causally---in fact, we will see that the pattern may even, counter-intuitively, suppress the rule. Preventing models from displaying these attention patterns can thus enhance, rather than reduce, the correlated output behavior.

Although our interpretations are not causally faithful, they predict the model's computation. Our results therefore support the notion that \textit{causal faithfulness} is distinct from our objective of \textit{holistic} interpretability as proposed in Sec. \ref{sec:philosophy}.

\subsubsection{Ablation method}

To test whether Dyck models rely on particular attention patterns, we measure accuracy after \emph{uniform attention ablation}---forcing every attention activation to attend equally to all prior tokens. This intervention preserves all other components but strips violation-tracking or other attention patterns. A model could, in principle, still apply \balanced{} after ablation: \citet{dyckcasestudy} showed uniform attention suffices to implement parentheses balancing. We apply the ablation to \emph{all} heads simultaneously, controlling for confounders from co-occurring head types. Note our results also hold for other ablation experiments ablating heads one at a time or using mean ablation (Appx.~\ref{apx:corrs}).

\subsubsection{Ablation results}\label{sec:ablation-results}

We find that certain types of violation-tracking attention---although correlated with high OOD accuracy---actually lower the OOD accuracy of the model when present (Fig.~\ref{fig:scatter_plot_multilayers}). When we ablate the attention of models with OOD \textit{violation-detecting} heads, we damage OOD accuracy, as might be expected if violation tracking is a key mechanism in implementing \balanced{}. By contrast,  when we ablate the attention of models with OOD \textit{sign-matching} heads, we actually \emph{improve} OOD accuracy. The latter type of hierarchical head, although just as correlated with hierarchical generalization as the former type, is not a key mechanism in the rule's implementation. Instead, it interferes with systematic hierarchical generalization.\footnote{Recall that some ID sign-matching heads become violation-detecting heads on OOD data.}
Ablating Main Auxiliary-Detecting heads in the question-formation setting does \emph{not} reverse the correlational signal: the ablation leaves OOD behavior essentially unchanged (Appx.~\ref{apx:qf-ablation}). Correlational predictive and causal analyses \emph{can} either converge or diverge; the predictive objective we propose is achievable in either case.

Our findings resist simple narratives about the mechanisms which implement a model's algorithm. Although all hierarchical heads are \textit{correlated} with hierarchical generalization, not all of them \textit{implement} that rule under causal tests. Therefore, even simple models can be complex enough to resist causal analysis, but correlational interpretations still provide predictive value. The decoupling is robust to the choice of ablation: \emph{mean} ablation and single-head ablation (Appx.~\ref{apx:corrs}) reproduce the same qualitative pattern as the uniform-attention ablation shown in Fig.~\ref{fig:scatter_plot_multilayers}---ablating sign-matching heads increases hierarchical OOD accuracy, while ablating violation-detecting heads decreases it.

Unlike the circuit which hypothetically implements \heuristic{}, OOD sign-matching  heads are \textit{promoted} by regularization, occurring more frequently when models are trained with weight decay (Appx. Fig.~\ref{fig:stacked_barplot_wd}). We therefore reject the position that sign-matching patterns are vestigial. Instead, we conjecture that these heads either develop as spandrels---side effects of learning \balanced{}---or they malfunction on OOD data. Module generalization failure is a ripe topic for future work.

\subsubsection{Causal roles are data dependent}\label{sec:datadependent} 

Interpretability research commonly tests proposed explanatory mechanisms by intervening on those mechanisms. We call this approach into question by demonstrating that a model can \textit{respond to an ablation differently on OOD and ID data}. Uniform attention ablation has substantial effects on OOD accuracy (Fig.~\ref{fig:scatter_plot_multilayers}), but leaves ID validation accuracy nearly unchanged, reduced by only 0.6\% on average (Appx. \ref{apx:head-ablation}); moreover, effects of ablation on ID and OOD accuracy are only weakly correlated ($\rho = 0.24$, $p<0.01$). These results suggest that a model might be able to compensate for the loss of a mechanism on ID data, while still heavily relying on the mechanism on OOD edge cases. Although a model may be applying the same rule ID and OOD at the output, the distribution shift reveals brittle elements of its implementation. We posit models could contain many redundant backup circuits that  compensate for ablation ID, but become unreliable under distribution shift. Without redundancy, model judgments become more sensitive to ablations on the remaining circuits. 

These results suggest that a negative ablation result is weak evidence against a proposed mechanism. Our findings mirror recent cases where current interpretability techniques failed to generalize on novel domains. Not only are SAE dictionary features highly data-dependent \citep{Paulo2025SparseAT}, but steering research shows features that are causal on one distribution may not be causal on new data 
\citep{kissane_saes_2024,kantamneni2025sparseautoencodersusefulcase,smith_negative_2025}. These concerns have spurred new interest in robust interpretability \citep{AlvarezMelis2018OnTR}.

\section{Discussion and Conclusions}
\label{sec:conclusion}

We show that interpretability can be used to predict model behavior on unseen inputs---even when the interpreted artifacts are not causally implicated in this behavior. 
Our findings suggest a new way of evaluating interpretations: by their ability to predict model behavior on unseen inputs. We intuit OOD generalization rules from ID representations and predict OOD generalization across two distinct settings: Dyck-1 classification and autoregressive English question formation. 

In particular, we find that some attention patterns causally support OOD generalization while others limit it. By conducting a \textit{correlational} analysis across a model population, we discover that violation-tracking attention patterns are predictive of hierarchical generalization behavior---even when causal analysis suggests they suppress this same behavior. These desiderata can be added to the current evaluation toolkit, which often focuses on in-distribution ablation response or correlation with data properties.

Causality is often held as the gold standard for mechanistic understanding. But causal findings are often compatible with multiple analyses that only expensive, precise ablations can distinguish, and proposed interpretations frequently fail to transfer to new settings~\citep{stander_grokking_2024, zhangnanda, makelov2023subspacelookingforinterpretability}. The literature on understanding attention has produced skepticism and fierce debate~\citep{bibal2022attention,dyckcasestudy} precisely because causal intervention has been treated as the lone arbiter of faithfulness. Our stance is correspondingly instrumentalist~\citep{duhem1954aim}: hierarchical structure is part of our scientific model of network behavior, sufficient if we can infer it from its traces. 

Predicting model behavior can be valuable independent of whether the underlying representational structure satisfies standard notions of mechanistic faithfulness. Even if perfect prediction remains elusive, flagging edge-case behavior could meaningfully improve evaluation and deployment. Overall, this work argues that the instrumental, \textit{algorithmic}-level goal of predicting model behavior is a worthwhile and tractable interpretability objective.

\section{Limitations}\label{sec:limitations}
While our work provides a proof-of-concept for predictive interpretability, it relies on a controlled synthetic setting with several simplifying assumptions. Our parentheses and question-formation tasks have a clear ground truth for generalization rules, and we train hundreds of models to analyze population-level patterns---conditions that may not hold when analyzing real-world models.

Future work could investigate whether mechanistic analyses can predict generalization in more realistic settings.
However, extending our approach faces challenges. First, similar analysis may be difficult in larger models because scale reduces---but does not eliminate---the impact of random variation and the isolated role of individual attention heads. Second, identifying interpretable structure is more difficult outside of synthetic settings with known OOD generalization behaviors.

Nevertheless, we believe representational geometry may provide clues about which capabilities a model has learned and how it may perform under distribution shift. Regardless of whether these geometric interpretations are considered faithful by standard metrics, predicting model behavior can be valuable independent of causal analyses. Even if perfect prediction remains elusive, flagging potential edge case behavior could meaningfully improve model evaluation and deployment decisions.

\newpage

\bibliography{custom}

\clearpage

\appendix

\onecolumn
\section{Glossary}
\label{apx:glossary}
\begin{table*}[!ht]
\centering
\small
\renewcommand{\arraystretch}{1.2}
\begin{tabular}{p{3cm} p{10.5cm}}\toprule
\textbf{Term} & \textbf{Definition and Usage} \\
\midrule

\textbf{Depth} & Parentheses sequence token characteristic. At index $j$, the token depth is $o(j) - c(j)$ where $o(j)$ and $c(j)$ are the cumulative counts of \open{} and \close{} up to $j$. 
\\ \midrule 

\textbf{Hierarchical \newline head} & Head that tracks a particular type of hierarchical attention pattern. Specifically, in Dyck, we evaluate if the \EOS{} attention weights persistently favor either violation or non-violation tokens (e.g., Equation \ref{eq:negative-depth} or \ref{eq:nonneg-depth}) on $\geq$80\% of sequences containing both violations and non-violations.

\vspace{0.5em}
This head type encompasses violation-detecting heads and sign-matching heads, and it is typically found in 2- and 3-layer models. \\ \midrule 

\textbf{EOS} & The end-of-sequence token where attention patterns are inspected. \\ \midrule 

\matched{} & Rule that a model can learn. \texttt{True} iff the number of open and close parentheses in a sequence are equal, regardless of their ordering (Equation \ref{eq:matched}). 

\vspace{0.5em}
\textbf{Marker}: Predicts \texttt{True} for all OOD inputs (0\% OOD accuracy). \\ \midrule 

\heuristic{} & Rule that a model can learn. Predicts \texttt{True} if first token is \open{}, \texttt{False} otherwise. This rule occurs in 1-layer models with 0 weight decay. 

\vspace{0.5em}\textbf{Marker}: $\sim$55\% OOD accuracy. \\ \midrule

\textbf{Depth-tracking \newline attention} & Dyck hierarchical attention pattern which reflects the tree structure of the preceding input sequence at an \EOS{} token by indicating which indices violate the condition in Equation \ref{eq:nested}. The pattern can favor either violation or non-violation tokens.  \\ \midrule

\textbf{ID data} & In-distribution training and validation data where negative examples satisfy neither the \matched{} nor \balanced{} rules. \\ \midrule

\balanced{} & Rule that a model can learn. Returns \texttt{True} iff the input is properly nested (i.e., satisfies both \matched{} and depth non-negativity, Equation \ref{eq:nested}). 

\vspace{0.5em}
\textbf{Marker}: Predicts \texttt{False} on all OOD examples (100\% OOD accuracy). \\ \midrule 

\textbf{Violation \newline token} & Token position that violates the \balanced{} condition in Equation \ref{eq:nested}, i.e., a position preceded by more close than open symbols. \\ \midrule 

\textbf{Violation-\newline detecting head} & Dyck hierarchical head that preferentially attends to tokens that violate Equation \ref{eq:nested} on $\geq 80$\% of sequences containing both violations and non-violations.

\vspace{0.5em} \textbf{Behavior}: Their presence is correlated with the hierarchical \balanced{} rule OOD, but ablating this head decreases  models' hierarchical behavior. \\ \midrule 

\textbf{OOD data} & Out-of-distribution test set. For Dyck, negative examples satisfy \matched{} but not \balanced{}. Higher accuracy is associated with \balanced{}. For question formation, OOD sentences begin with different auxiliaries for linear vs hierarchical rules. Higher accuracy is associated with hierarchical behavior. \\ \midrule 

\textbf{Sign-matching head} & Hierarchical head that attends to violation tokens if the last token violates Equation \ref{eq:nested} and non-violation tokens if the last token does not violate Equation \ref{eq:nested}, on $\geq80$\% of sequences containing both violations and non-violations.

\vspace{0.5em}\textbf{Behavior}: Their presence is correlated with the hierarchical \balanced{} rule OOD, but ablating this head increases models' hierarchical behavior. \\ \midrule 

\textbf{Main Auxiliary-\newline Detecting head} & Question formation analog of a Dyck hierarchical head. At the question-formation readout position ({\color{darkblue}\texttt{quest}}), a head in one of the first three layers of the model places attention weight $> 0.2$ on the matrix-auxiliary token on $\geq 80$\% of ID sequences.

\vspace{0.5em}\textbf{Behavior}: Models with such ID heads correlate with higher OOD (hier.) accuracy. \\ \midrule

\textbf{Uniform attention \newline ablation} & Flattens attention to a uniform weight distribution. Used to test causal importance of particular attention heads and patterns. \\ \midrule 

\textbf{Vestigial circuit} & A sub-circuit used early in training, but not necessary for model performance at the end (e.g., \heuristic{} circuit in unregularized 1-layer models). At the end of training, such circuits may be associated with rules applied OOD but not ID. \\ 

\bottomrule
\end{tabular}\vspace{1.2em}
\caption{Glossary of key terms used in this paper.}
\label{tab:glossary}
\end{table*}

\twocolumn 

\newpage

\section{Attention Activation Examples}\label{apx:attn-examples}

We include a figure to clarify what we mean by violation and non-violation tokens. 
\begin{figure}[!ht]
    \centering
    \includegraphics[width=\linewidth]{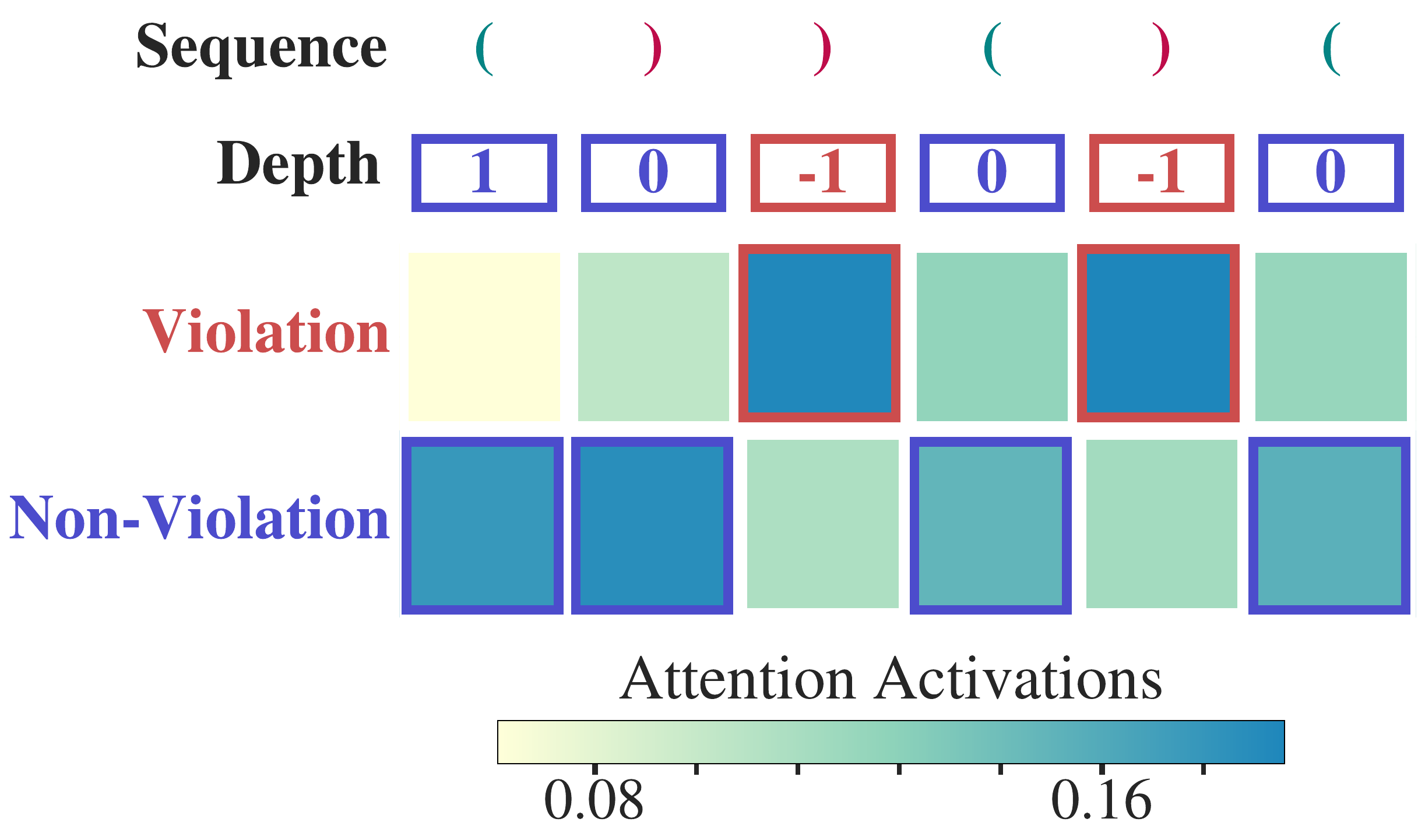}
    \caption{Examples of attention patterns favoring violation (top row) and non-violation (bottom row) tokens at the \EOS{} position. The input sequence is shown along the top; each token's tree depth $o(j) - c(j)$ is displayed below. Tokens with depth $<0$ are violations of Equation \ref{eq:nested}.}
    \label{fig:attn_examples_appx}
\end{figure}

\section{Reproducibility Statement}\label{apx:reproducibility}

All 270 models, data, and code in the Dyck setting are freely available for use without restriction at 
\href{https://github.com/vli31/id-predict-ood}{https://github.com/vli31/id-predict-ood}.
Hyperparameter settings for each run are extensively documented in the main text and appendix of this paper. We recommend that future work experiment further with increasing scale and study the particular interactions between heads, to extend our descriptions of representation to be more granular.

We adapted the minGPT software artifact which has an MIT license \cite{minGPT}. Packages used for evaluation include SciPy (version 1.14.0). To train models for the question-formation setting, we used the open source output of \citet{qin2024itreedatadrives} available at \href{https://github.com/sunnytqin/hier_gen}{https://github.com/sunnytqin/hier\_gen}.

\section{Dyck Dataset Generation Details}\label{apx:dataset}

We can think of the sequence length distribution as though we are generating \balanced{} trees with a 50\% probability of recursing at each node, but discarding identical sequences. Each sequence of symbols, sampled uniformly at random, is then sorted according to which rule it follows.

\subsection{Dataset parentheses sampling}

We create datasets as follows: 
\begin{enumerate}
    \item Sample an even sequence length $n = 2 + 2B$ with $B \sim \text{Binomial}(19, 0.5)$, so that $n$ ranges over even values from $2$ to $40$. These properties ensure that our samples are concentrated around a reasonable center, reducing extreme sequence lengths that could occur with other distributions like the Uniform. Also note that our maximum sequence length is $40$. 
    \item Generate a uniformly random parentheses sequence of length $n$ with the desired attributes.
    \begin{itemize}
        \item To generate a uniformly random sequence that is neither \matched{} nor \balanced{}, we choose each character independently from the set \{ \open{}, \close{} \}. If the resulting sequence satisfies \matched{}, we discard it.
        \item To generate a uniformly random sequence that is \matched{} but not \balanced{}, we randomly permute $n/2$ \open{} parentheses and $n/2$ \close{} parentheses. If the resulting sequence is \balanced{}, we discard it and generate a new one.
        \item To generate a uniformly random sequence that is \balanced{}, we use the algorithm of {\citet{arnoldsleep}}. 
    \end{itemize}
    \item If the sequence generated does \emph{not} already appear in the dataset, add it to the dataset.
\end{enumerate}
Thus, each length-$n$ sequence $s$ with the desired attributes is equally likely to be chosen, and it is chosen at most once. Since we discard repeats, the empirical distribution of sequence lengths is skewed towards longer sequences, as short sequences are likely to be repeated. 

We tokenize the \open{} and \close{} characters in addition to start, end, and padding tokens (\texttt{BOS}, \texttt{EOS} and \texttt{PAD}, respectively, with \texttt{PAD} appended to the end of the sequence) to ensure each sequence for classification has length $42$ (including start and end tokens). In other terms, we create a sequence of form:
$$s_0s_1\ldots s_n \ldots s_{41},$$
where $s_0$ is the beginning-of-sequence token \texttt{BOS}, $s_1$ through $s_{n}$ make up the $n$-length parentheses sequence $s$, $s_{n+1}$ is the end-of-sequence \texttt{EOS} token, and $s_{n+2}$ through $s_{41}$ are \texttt{PAD} tokens.

Our ID datapoints are randomly split into training and validation datasets. Each ID set contains the same number of \texttt{True} examples (following both \matched{} and \balanced{}) and \texttt{False} examples (following neither \matched{} nor \balanced{}). Our OOD test set consists of parentheses sequences which follow \matched{} but not \balanced{}, i.e., sequences with the same number of open and close parentheses, but in which the parentheses are not properly nested (e.g., $ \close{}\close{}\open{}\open{} $). See Table \ref{tab:firstsymbol} for examples.

Empirically, we find some models classify sequences by a \heuristic{}  heuristic OOD. These models check whether $s_1 =  \open $ and label an OOD sequence as \texttt{True} if the first character is $\open$ and \texttt{False} if it is $\close$. (In-distribution, no models follow \heuristic{}, which would fail to achieve full accuracy ID.)

\begin{table*}[ht]
\centering
\caption{The \matched{} and \balanced{} rules applied to example parentheses sequences in our ID and OOD test sets. Also, classifications of the same parentheses sequences according to the OOD \heuristic{} heuristic. Notice that this rule does not achieve perfect accuracy ID.}\label{tab:firstsymbol}
\begin{tabular}{@{}ccc|ccc@{}}
\toprule
Dataset              & \matched                        & \balanced                       & Possible $s_1$ & Example                                & \heuristic{}   \\ \midrule
ID                   & \texttt{True}                   & \texttt{True}                   & \open{}                & \open{}\close{}\open{}\open{}\close{}\close{} & \texttt{True}  \\ \midrule
\multirow{2}{*}{ID}  & \multirow{2}{*}{\texttt{False}} & \multirow{2}{*}{\texttt{False}} & \open{}                & \open{}\open{}\close{}\open{}\close{}                & \texttt{True} \\
                     &                                 &                                 & \hspace{3pt}\close{}               &                                \close{}\open{}\open{}\open{}\close{}               & \texttt{False}  \\ \midrule
\multirow{2}{*}{OOD} & \multirow{2}{*}{\texttt{True}}  & \multirow{2}{*}{\texttt{False}} & \open{}                & \open{}\close{}\close{}\open{}                & \texttt{True}  \\
                     &                                 &                                 & \hspace{3pt}\close{}               &                  
                     \close{}\open{}\close{}\open{}
                     
                     & \texttt{False} \\ \midrule
---                   & \texttt{False}                  & \texttt{True}                   & ---                     & Does Not Exist                                           & ---             \\ \bottomrule
\end{tabular}\vspace{1em}
\end{table*}

\subsection{Random data order implementation}

Our train set contains $200000$ distinct datapoints. During training, we repeat this set five times, so all models were exposed to $1$ million total parentheses sequences (including $5$ repeats of each) across the course of training. The random seed used for data ordering, or ``shuffle seed,'' determines the order of data within each block of $200000$ training examples, so during training, a single model is exposed to the same examples in five different orderings. Models with the same shuffle seed hyperparameter encounter the 1 million total training datapoints in exactly the same order (data within each block is shuffled in a consistent way).

\section{Factors in rule selection}\label{sec:factors}

For consistency, we define OOD accuracy with respect to the \balanced{} rule. Thus a model achieving 100\% OOD accuracy classifies each \textsc{Non-Nested}, \matched{} sequence in our OOD test set as \texttt{False}. Correspondingly, models with 0\% OOD accuracy learn \matched{}, classifying every OOD example as \texttt{True} (Table \ref{tab:firstsymbol}).

\subsection{Architecture}\label{sec:architecture}

\begin{figure*}[ht]
    \centering
    \begin{subfigure}{0.47\textwidth}
    \includegraphics[width=\linewidth]{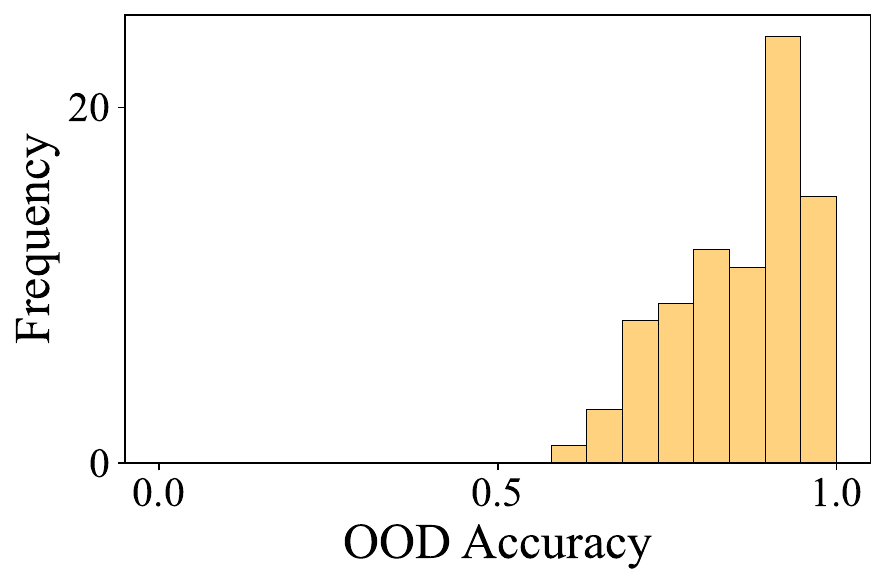}
        \caption{OOD-test accuracy for LSTMs.}
    \end{subfigure}
    \hfill
    \begin{subfigure}{0.47\textwidth}
    \includegraphics[width=\linewidth]{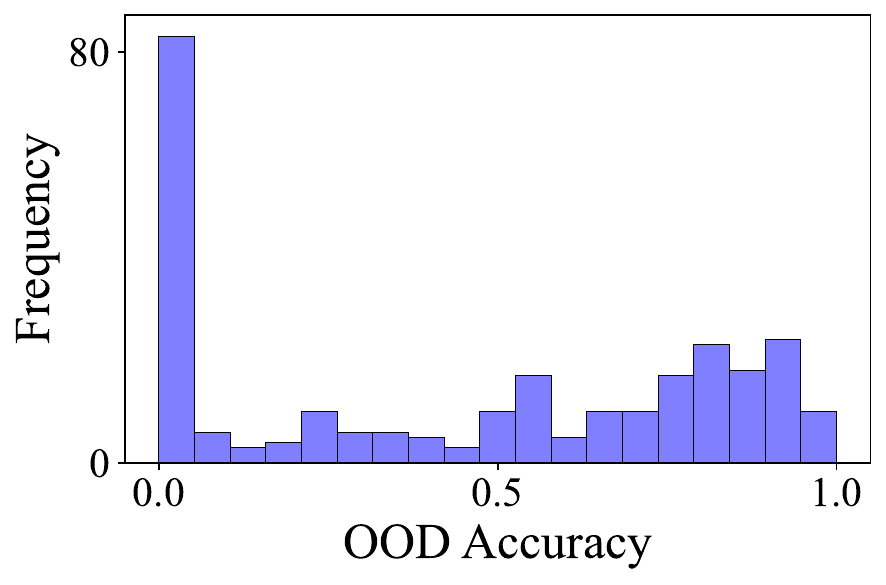}
        \caption{OOD-test accuracy for Transformers.}
    \end{subfigure}
    \caption{Last OOD test accuracy for LSTM and Transformer models that achieve 99\%+ ID accuracy. When LSTMs converge to near-perfect ID accuracy, they consistently also apply the \balanced{} rule to OOD data. Transformers, meanwhile, apply a variety of rules and exemplar-based behavior OOD.}
    \label{fig:lstm}
\end{figure*}

By comparing Transformers to LSTMs, we confirm existing findings \citep{abnar2020transferringinductivebiasesknowledge,mccoy-etal-2020-syntax,tran2018importancerecurrentmodelinghierarchical, saphra-lopez-2020-lstms} that LSTMs are intrinsically hierarchical while Transformers are not. The inductive bias of the LSTM architecture places every trained model at more than 60\% accuracy on the OOD generalization set, indicating that none of these models learn \matched{} and all are closer to the hierarchical \balanced{} rule. In contrast, Transformer models exhibit an OOD accuracy distribution with two peaks: one near 0\% (indicating perfect application of the \matched{} rule) and a smaller one at 90\% (indicating a tendency towards \balanced{}). Overall, 24.4\% of Transformers learn \matched{} perfectly, achieving zero OOD accuracy at the last step in training (Figure \ref{fig:lstm}).

These architectures ranged in parameter count with 1-, 2-, and 3-layer Transformers having 53K, 103K, and 153K parameters, respectively, and LSTMs around 250K parameters. Overall, these models took around 25 H100 GPU hours to train.

\subsection{Width}\label{sec:model_width}

\begin{figure}[!ht]
    \centering
        \includegraphics[width=\linewidth]{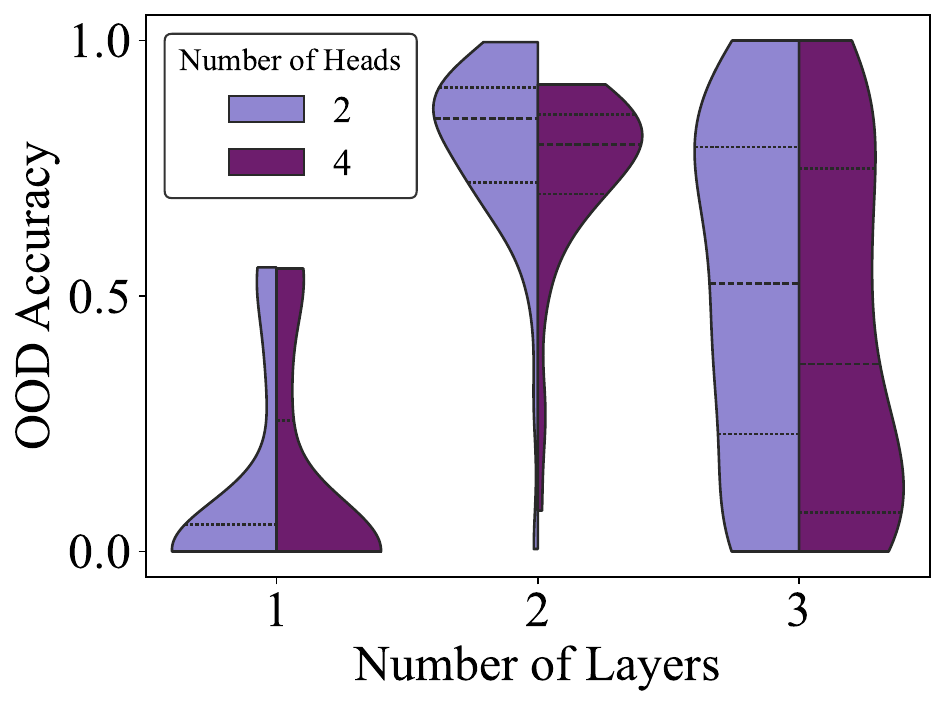}
    \caption{Unlike weight decay and depth (Figure \ref{fig:wd}), width is not a substantial factor in final OOD rule selection in this setting. The Mann--Whitney U test finds no statistically significant differences in the distribution of last OOD accuracy over width (all $p > 0.05$).}
        \label{fig:width}
\end{figure}

\begin{figure}
\centering

\includegraphics[width=0.96\linewidth]{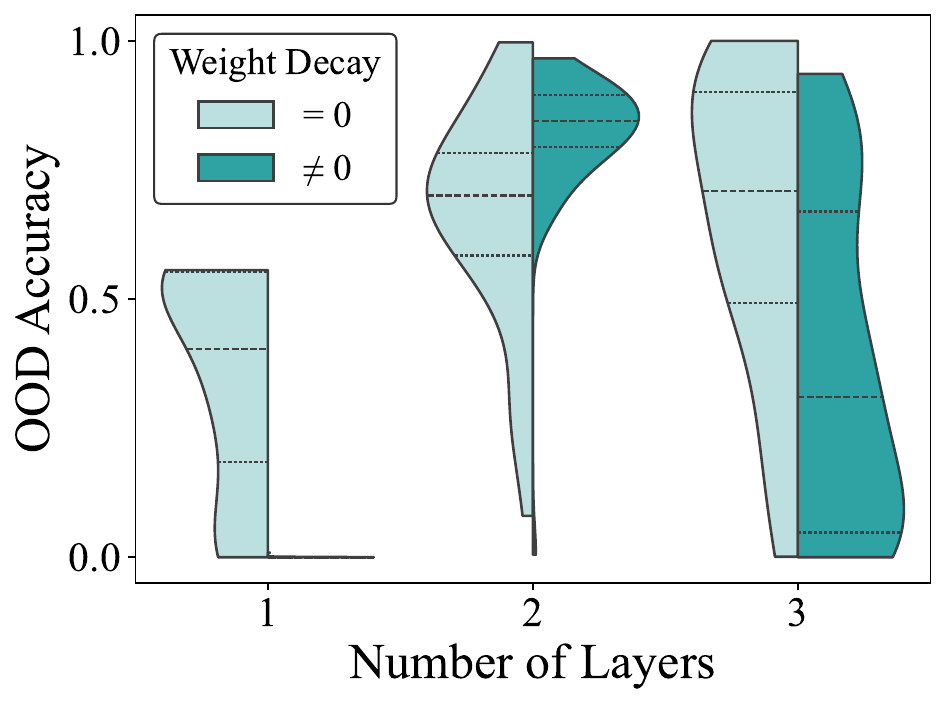}
    \caption{Weight decay and depth substantially impact OOD rule choice. The Mann--Whitney U test finds significant differences in OOD accuracy distribution between the absence and presence of weight decay across depths ($p \ll 0.01$ across layers).}
    \label{fig:wd}
\end{figure}

Using the non-parametric Mann--Whitney U test to detect differences between distributions, we find that the number of Transformer heads has no significant effect on the distribution of OOD accuracies for any depth of model (Figure \ref{fig:width}). This result, which we show holds over randomness in model initialization and data exposure order, adds to a growing body of evidence across settings that changing Transformer width has little effect on model expressivity and OOD generalization \citep{petty2024, tay2022scale}.

\subsection{Depth}\label{sec:model_depth}

\begin{figure*}[!ht]
    \centering
    \includegraphics[width=0.62\linewidth]{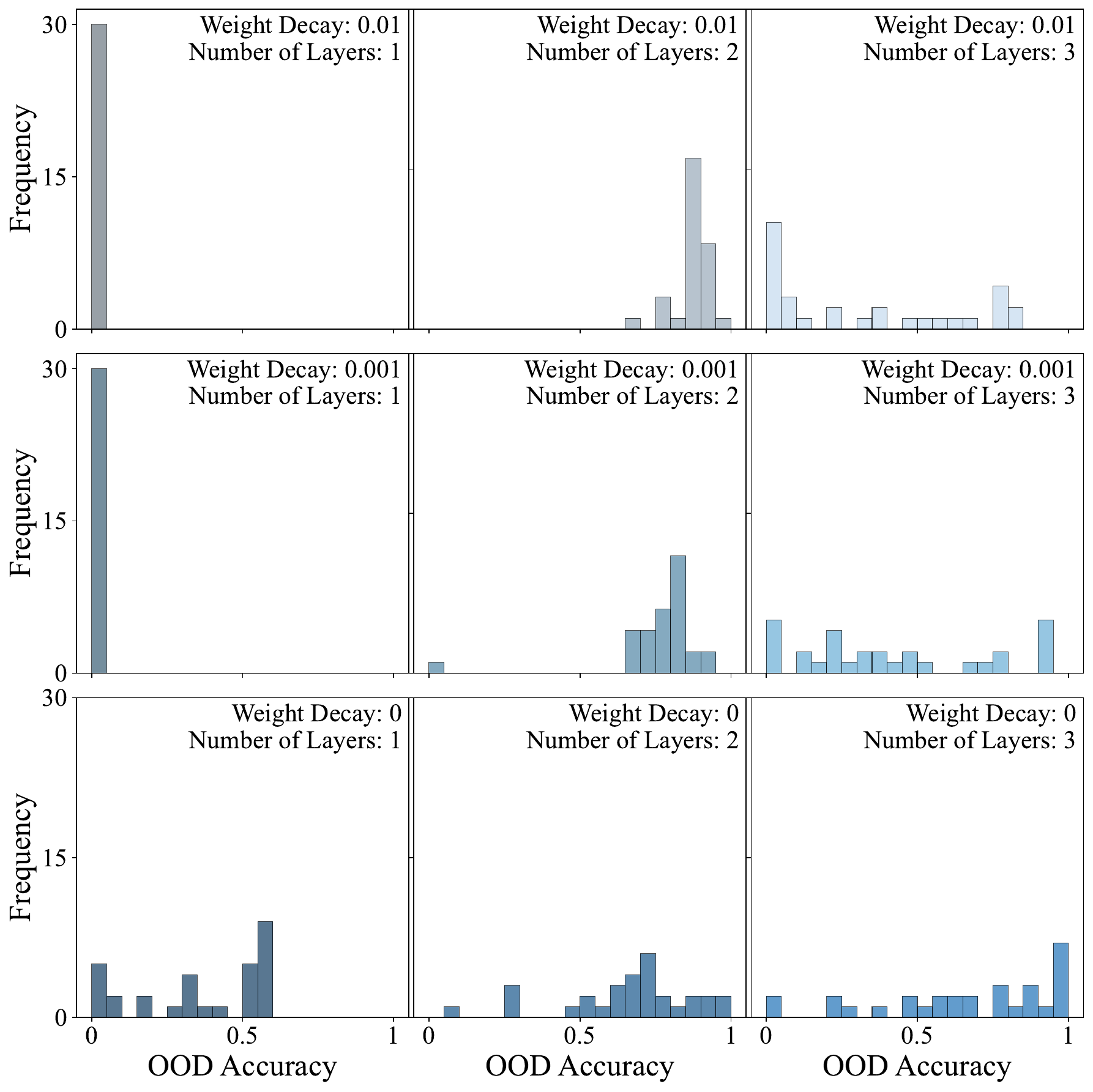}
    \caption{Final accuracy on OOD-test for Transformers of varying depths and weight decays. 1-layer Transformers learn \matched{} or \heuristic{}. Deeper Transformers can learn  \matched{}  or approximate \balanced{}, with 2-layer Transformers most likely to learn \balanced{} and 3-layer Transformers instead exhibiting more complex OOD generalization behavior.}
    \label{fig:depth}
\end{figure*}

In this setting, depth---unlike width---is a significant factor in rule selection, determining the peaks of the OOD behavior distribution. For instance, among the 66 out of 270 Transformers (24\%) which achieve perfect \matched{} behavior with 0\% OOD accuracy (Figure \ref{fig:lstm}), there are $61$ 1-layer models, no 2-layer models, and $5$ 3-layer models. (Note our overall model population is evenly split into 90 each of 1-, 2-, and 3-layer models.)

We find that 1-layer models fall into a bimodal distribution centering on two rules: \matched{} (characterized by 0\% OOD accuracy) and \heuristic{} (which gives $\sim$55\% OOD accuracy). The 18 models that generalize according to \heuristic{} all exhibit nearly identical judgments on specific examples, following a rule associated with returning a \texttt{True} label if the input begins with \open{}. Because we only consider models with at least 99\% validation accuracy, the models in question must also have other subroutines that are successfully applied ID but dominated by \heuristic{} OOD.

 \begin{figure}[ht]
    \centering
    \includegraphics[width=0.5\textwidth]{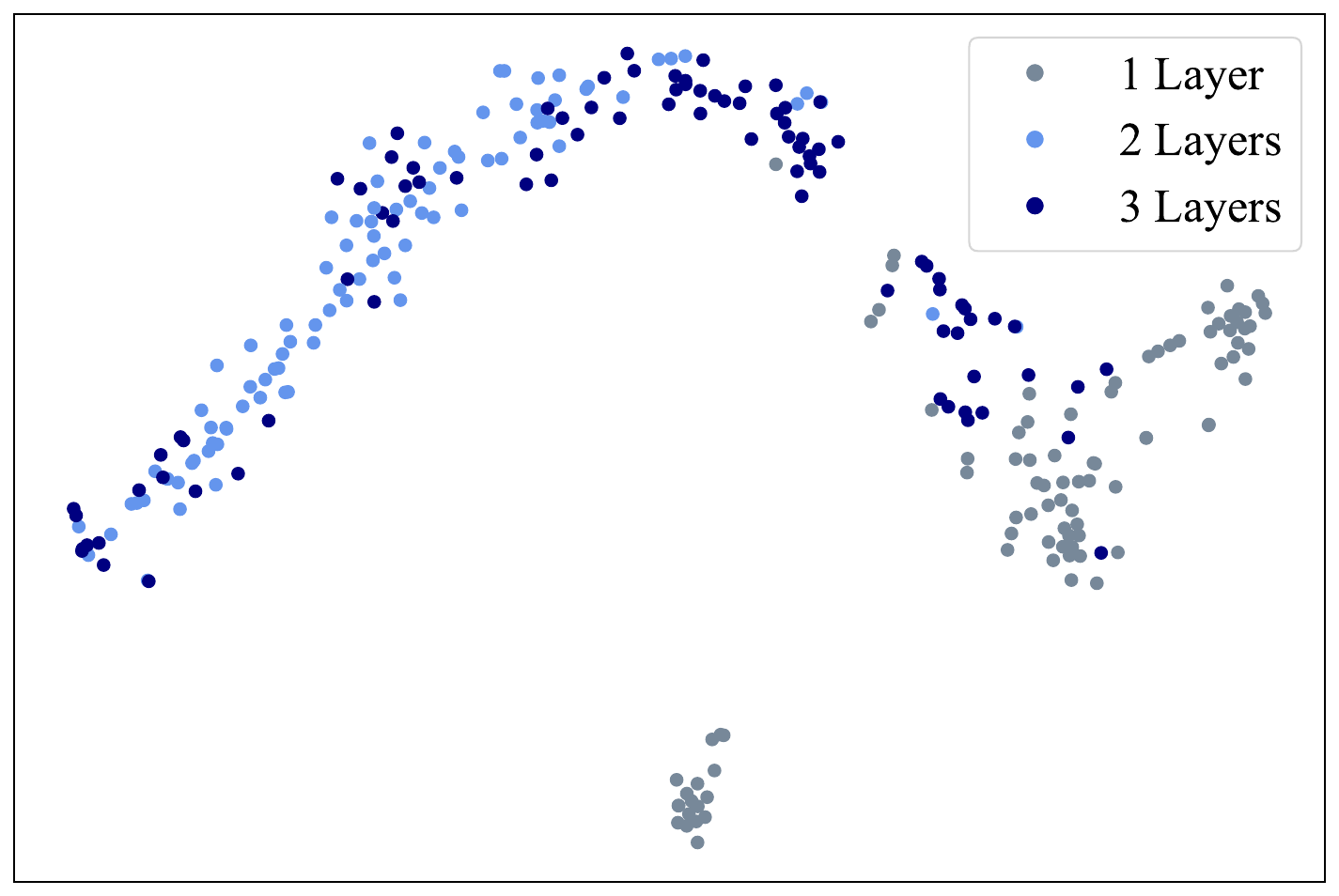}
    \caption{t-SNE of models' final OOD classifications colored by model depth.}
    \label{fig:clusters}
\end{figure}

We group models via t-SNE (perplexity 12) based on their judgments on the OOD test set (Figures \ref{fig:tsne-ood-prob} and \ref{fig:clusters}). \heuristic{} forms an outlier cluster in model judgments of $1$-layer models, which otherwise  learn \matched{}.

By contrast, only 2\% of 2-layer models are \matched{}-leaning (determined by $< 20\%$ accuracy OOD) and none learn \heuristic{}. The mode of this model distribution is instead at 90\% accuracy, firmly suggesting that most 2-layer models have approximately learned \balanced{}. Among 3-layer models, behavior varies enormously, with the distributional mode defined by the 20 models with $<0.1$ OOD accuracy which learn \matched{}. Across all 270 training runs, 10.4\% of models achieve at least 90\% accuracy, including $15$ 2-layer and $13$ 3-layer models.

\subsection{Regularization}\label{sec:regularization}

Weight decay has a significant impact on the distribution of rules models learn. Without any weight decay, models that generalize ID can converge on a variety of OOD generalization rules with wide distributions. With weight decay, particularly among smaller models, models have more similar OOD behaviors. For example, while 2-layer Transformers show a consistent tendency to prefer \balanced{}, they achieve higher OOD accuracy more reliably when weight decay is applied (Figure \ref{fig:depth}, Figure \ref{fig:wd}).

Among 1-layer models, training with weight decay always results in convergence to the \matched{} rule. Among 1-layer models trained without weight decay, 60\% (18 of 30) converge to \heuristic{} with $\sim$55\% accuracy on the OOD test set (Figure \ref{fig:wd}). The presence of all \heuristic{}-learning models in 1-layer models with weight decay $0$ indicates regularization can help prune away vestigial model features unnecessary for ID generalization. The presence of circuits supporting \heuristic{} may not impact ID performance, but in the absence of regularization, such features significantly decrease OOD performance.

\subsection{Randomization in Weight Initialization and Data Order}

\begin{figure*}[!ht]
    \centering
    \includegraphics[width=0.66\linewidth]{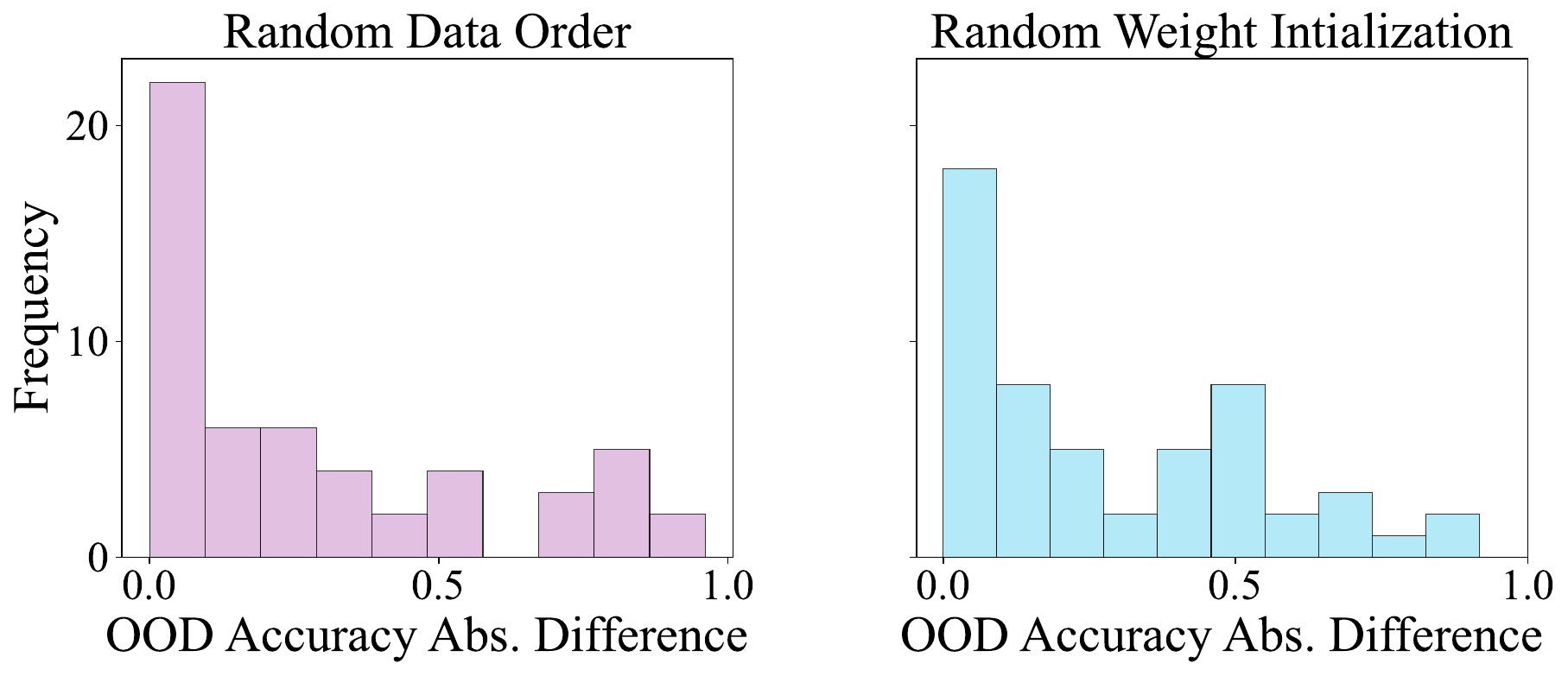}  
    \caption{Data order (left) and weight initialization (right) are equally influential random factors in OOD behavior, as shown by the distribution of the ranges of final OOD accuracy between models trained with differing data ordering and weight initialization.}
    \label{fig:rdmness}
\end{figure*}

Existing work \cite{qin2024itreedatadrives,chan2022datadistributionalpropertiesdrive, zhao2024distscaling, junejaetal} has investigated the impact of random weight initialization and data order on model performance. \citet{dodge2020finetuningpretrainedlanguagemodels} varied these factors in BERT fine-tuning, finding that modifying either factor had significant, comparable impacts on model performance. We investigate the impact of the two sources of random variation that account for differing model behaviors, and find that while they do not affect ID accuracy, both dataset ordering and model initialization affect generalization behavior.

Since we train models across three data shuffle seeds and five model random seeds, for a fair comparison between ranges of the impact of these factors on final OOD accuracy, we randomly select three random seeds to plot and compare the ranges of performance across shared hyperparameter conditions.

In Figure \ref{fig:rdmness}, we show that in a plurality of models trained, OOD performance was impacted by at least 10\%, with the maximum difference due to either one of the two random factors reaching an above 90\% difference in OOD behavior. The similar distribution in the impact of random initialization and data order is aligned with previous work and indicates both factors are important to determining model OOD performance and should be accounted for in building robust ML systems.

\section{Training dynamics of different rules}\label{sec:trdy}

\begin{figure*}[!ht]
    \centering
    \includegraphics[width=0.85\textwidth]{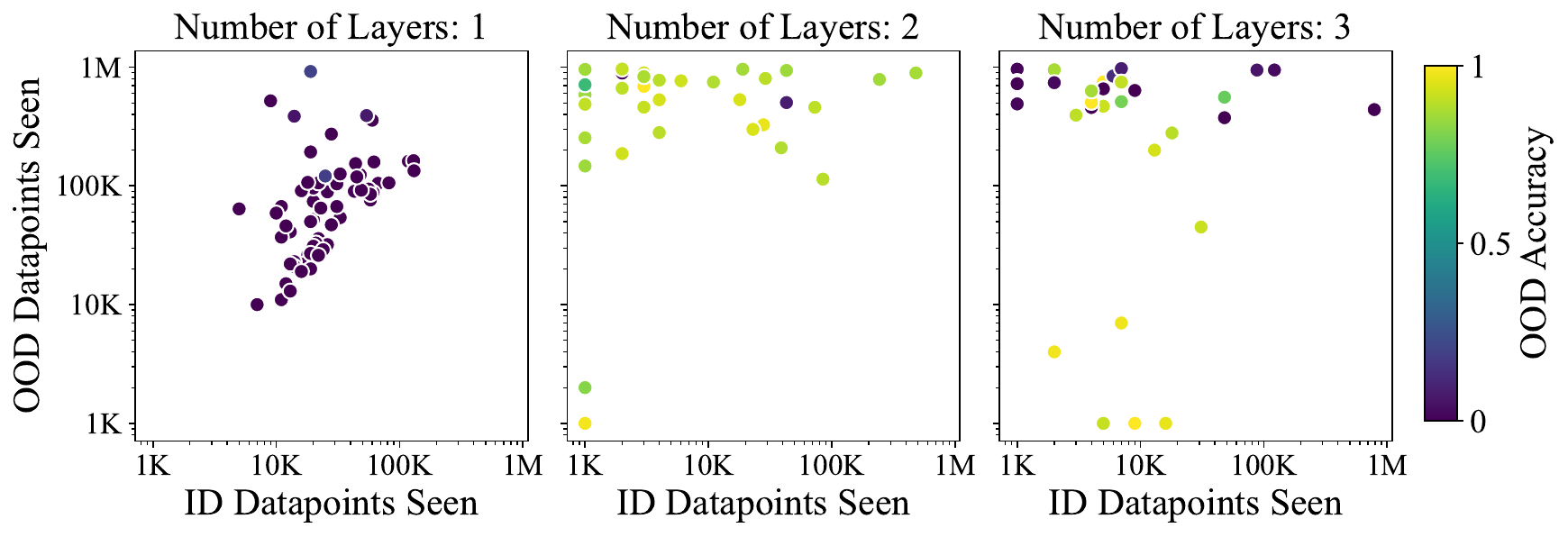}
    \caption{Illustration of generalization rules across training. ID convergence occurs when models achieve $\geq$ 99\% ID accuracy for $>$ 99\% of remaining datapoints seen---all Transformers achieve this metric after 900K datapoints. We define OOD convergence to either \matched{} or \balanced{} as Transformers achieving $\leq0.2$ or $\geq0.8$ accuracy for $>$ 99\% of the rest of the model run, respectively, after seeing at most 975K datapoints---53\% of Transformers achieve this metric. Using these metrics, this plot shows the number of datapoints seen before OOD and ID convergence, excluding models that do not converge OOD.}
    \label{fig:training-convergence}
\end{figure*}

\begin{figure*}[!ht]
    \centering

    \begin{subfigure}{.25\linewidth}
    \includegraphics[width=\linewidth]{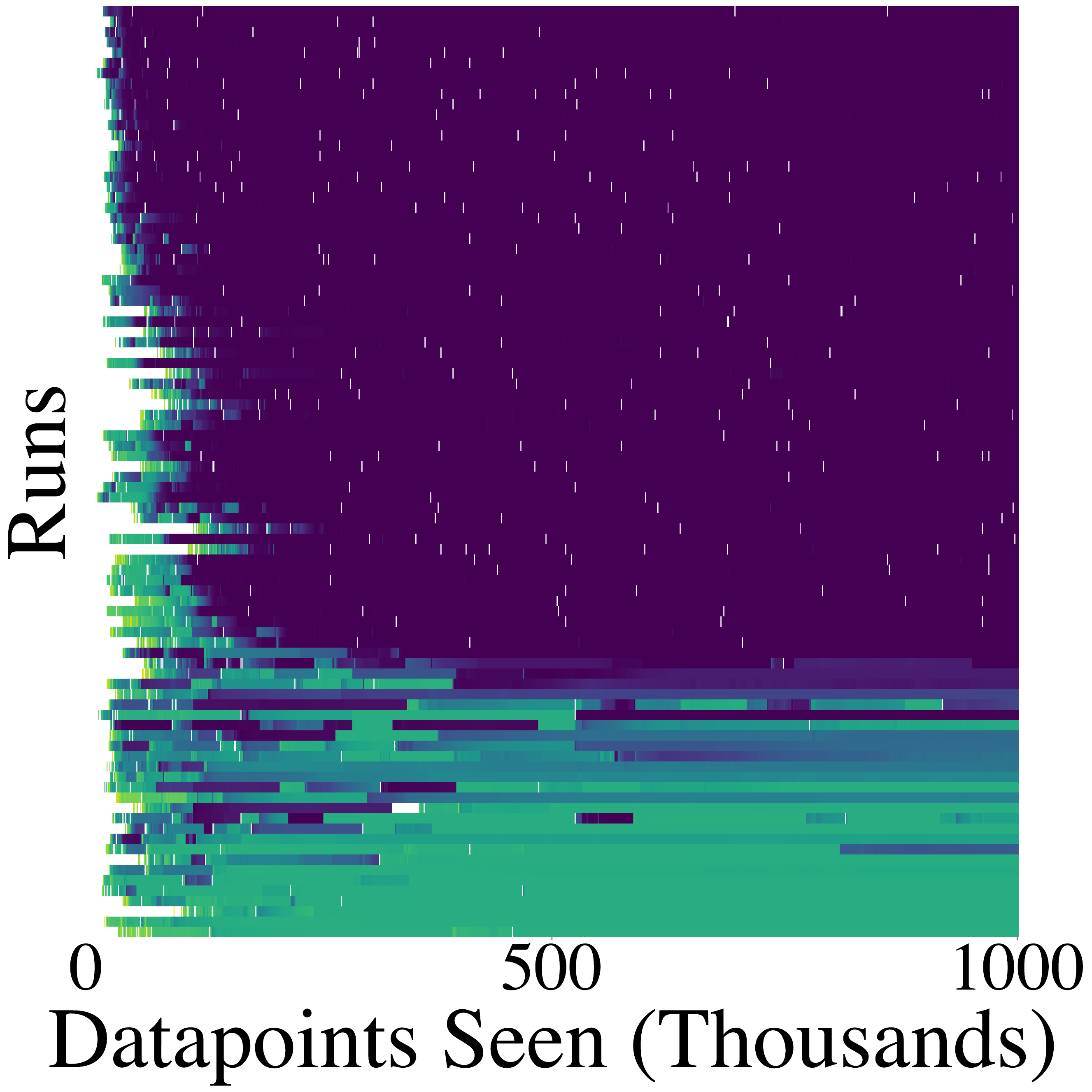}
      \caption{1-layer Transformers}
      \label{fig:mean_loss_tok_1}
    \end{subfigure}
    \hfill
    \begin{subfigure}{.25\linewidth}
    \includegraphics[width=\linewidth]{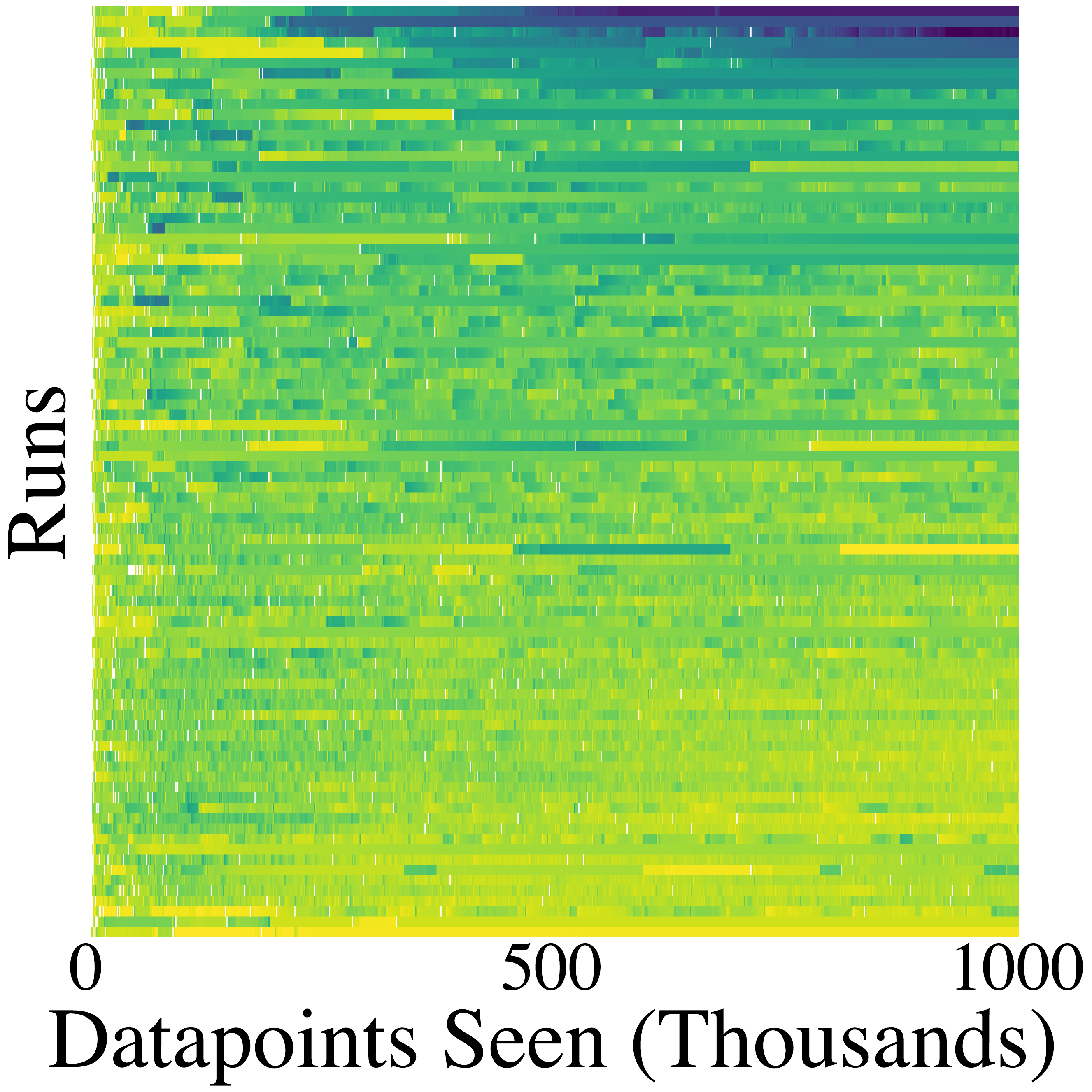}
      \caption{2-layer Transformers}
      \label{fig:mean_loss_tok_2}
    \end{subfigure}
    \hfill
    \begin{subfigure}{.25\linewidth}
    \includegraphics[width=\linewidth]{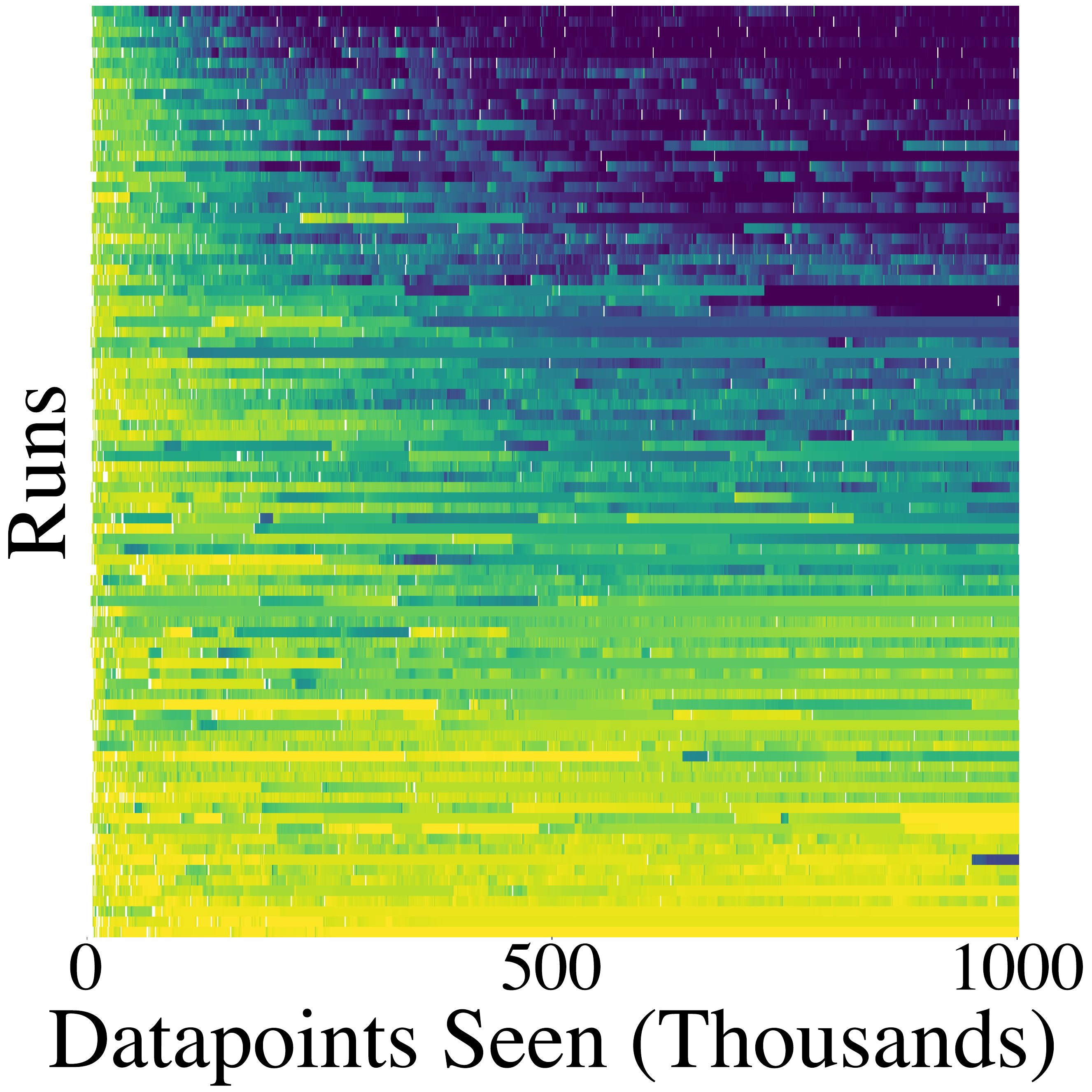}
      \caption{3-layer Transformers}
      \label{fig:mean_loss_tok_3}
    \end{subfigure}
    \caption{Heatmaps showing model training dynamics broken down by depth, where purple and yellow indicate model adherence to the \matched{} and \balanced{} rule, respectively (using the same scale as Figures \ref{fig:tsne-ood-prob}, \ref{fig:heuristic-phase}, and \ref{fig:training-convergence}). Colored cells indicate the OOD accuracy of a particular run when ID accuracy is at least $0.99$.}
    \label{fig:heatmap-layer}
\end{figure*}

Some OOD generalization rules can converge simultaneously with ID performance, whereas others take longer to learn after the model successfully learns ID. In our setting, models can acquire and stabilize into the \matched{} rule, but generally take longer to converge to the \balanced{} rule. 

Although overall, models tend to classify OOD sequences as \texttt{False} at the outset of training---likely because the \texttt{False} training examples, being sampled uniformly at random, are far more diverse---they rarely stabilize immediately at high rates of \texttt{False} (i.e., equivalent to a \balanced{} rule). The models that stabilize at a \balanced{} rule, as seen in Figure \ref{fig:training-convergence}, often stabilize long after ID convergence. In other words, we see an example of \textit{structural grokking} \citep{structuralgrokking}. 
These results support the idea that \balanced{} is a more difficult rule to fully learn. Indeed, only three Transformer models adhere completely to the \balanced{} rule by classifying all OOD examples as \texttt{False}. The training dynamics of different rules broken down in detail by model depth are also shown in Figure \ref{fig:heatmap-layer}. Notably, deeper models have higher final OOD accuracy variance and also display greater variance during training, possibly because of their higher expressivity.

\subsection{Evaluating the Heuristic}

The \heuristic{} heuristic, in contrast to the \balanced{} and \matched{} rules, is not used by any models ID, where it would produce a low accuracy. However, it is used by some 1-layer models OOD, and it produces an OOD accuracy of $ \approx 0.55$, reflecting the fact that around 55\% of our OOD test examples begin with close parentheses.

As discussed in Section \ref{sec:heuristic}, a majority of 1-layer models appear to pass through a \heuristic{} heuristic phase, although this heuristic does not persist until the end of training among models trained with weight decay. In this case, we say the models ``appear'' to pass through this phase, rather than asserting that they do, because we are only able to verify their behavior on individual datapoints at our five saved model checkpoints. However, at those model checkpoints, we are able to confirm that 1-layer models whose OOD accuracy is between $0.54$ and $0.56$ indeed almost always make OOD judgments based on first symbol. We therefore posit that throughout training, instances of OOD accuracy in this range likely reflect the \heuristic{} heuristic, though we cannot rule out that some such instances reflect some other heuristic which coincidentally produces the same OOD accuracy. Notably, 2- and 3-layer models also reach accuracies in this range while training, but, checking them at saved checkpoints, we find they do not learn \heuristic{}.

\section{Attention Heads Classified by OOD Behavior}

 \begin{figure*}[!ht] 
\begin{center}
\includegraphics[width=0.6\textwidth]{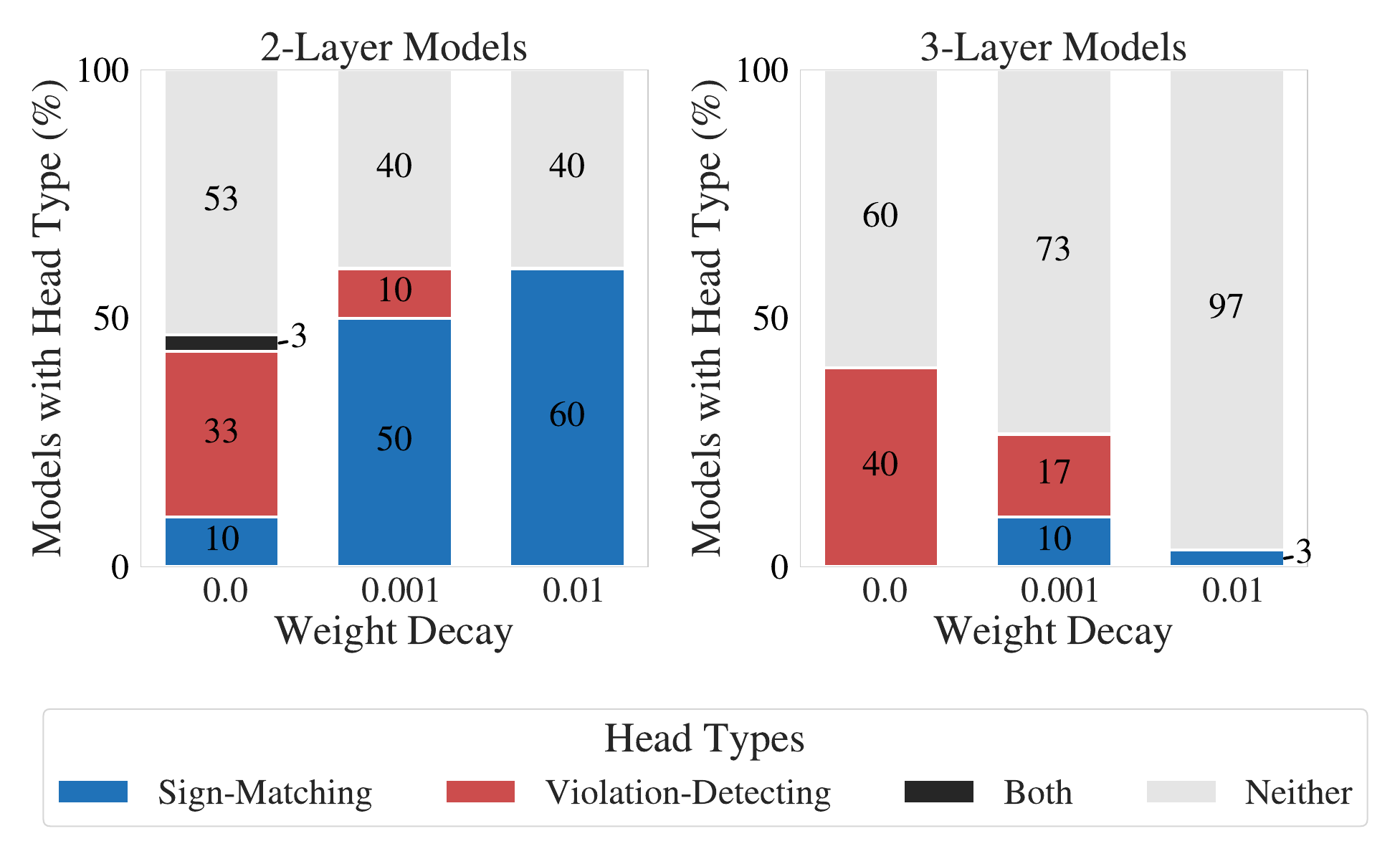}
\end{center}
\caption{Percentage of 2- and 3-layer models containing each head type, by weight decay. Head types are classified according to their OOD behavior.}\label{fig:stacked_barplot_wd} 
\end{figure*}

We break down the presence of types of head across 2- and 3-layer models (Figure \ref{fig:stacked_barplot_wd}).

In Section \ref{sec:predicting-ood}, we showed that classifying hierarchical heads according to their behavior on the ID validation set is predictive of generalizing according to the \balanced{} rule. It is also possible to conduct the same analysis using behavior on the OOD test set. We find that the presence of OOD hierarchical heads is similarly predictive of generalizing according to the \balanced{} rule (Appx.~\ref{apx:classifier}). All types of hierarchical head appear to correlate similarly strongly with \balanced{} generalization.

\section{ID Hierarchical Heads as a Classifier}\label{apx:classifier}

We can recast our predictive interpretability claim as a concrete classifier. For Dyck, ``model has at least one ID hierarchical head'' is a binary prediction of ``model implements \balanced{} on OOD.'' Across the 2- and 3-layer Dyck population ($n=180$), thresholding the true label at OOD accuracy $> 0.5$ yields \textbf{precision $= 0.96$, recall $= 0.68$} (F1 $= 0.79$); the OOD-classified head detector reaches precision $0.94$, recall $0.53$ (F1 $= 0.68$). For the question-formation setting ($n=79$), the Main Auxiliary-Detecting head detector reaches precision $0.76$, recall $0.85$ (F1 $= 0.80$) at the same OOD-acc cutoff. The two settings sit at opposite ends of the precision/recall trade-off: in Dyck the head detector is \emph{high-precision, lower-recall}---models flagged as carrying a hierarchical head almost always generalize hierarchically OOD, while some hierarchical-OOD models implement the rule without a detectable head (consistent with \citet{dyckcasestudy}'s finding that hierarchical behavior can also arise through nearly uniform attention). In question formation the Main Auxiliary-Detecting head is instead \emph{high-recall}: most hierarchically-generalizing models carry it, at moderate precision. In both settings, carrying the head is a reliable positive signal for hierarchical OOD behavior.

\subsection{Held-out evaluation}\label{apx:heldout}
In the previous section, we fit precision and recall on the population. We also evaluate, on unseen models, predictors of whether a model lies in the \balanced{} region of Fig.~\ref{fig:tsne-dyck}. This predictor sees only ID information. We report AUROC, the probability that a randomly chosen \balanced{} model is ranked above a randomly chosen non-\balanced{} model.

\begin{figure}[!ht]
\begin{center}
\includegraphics[width=0.8\linewidth]{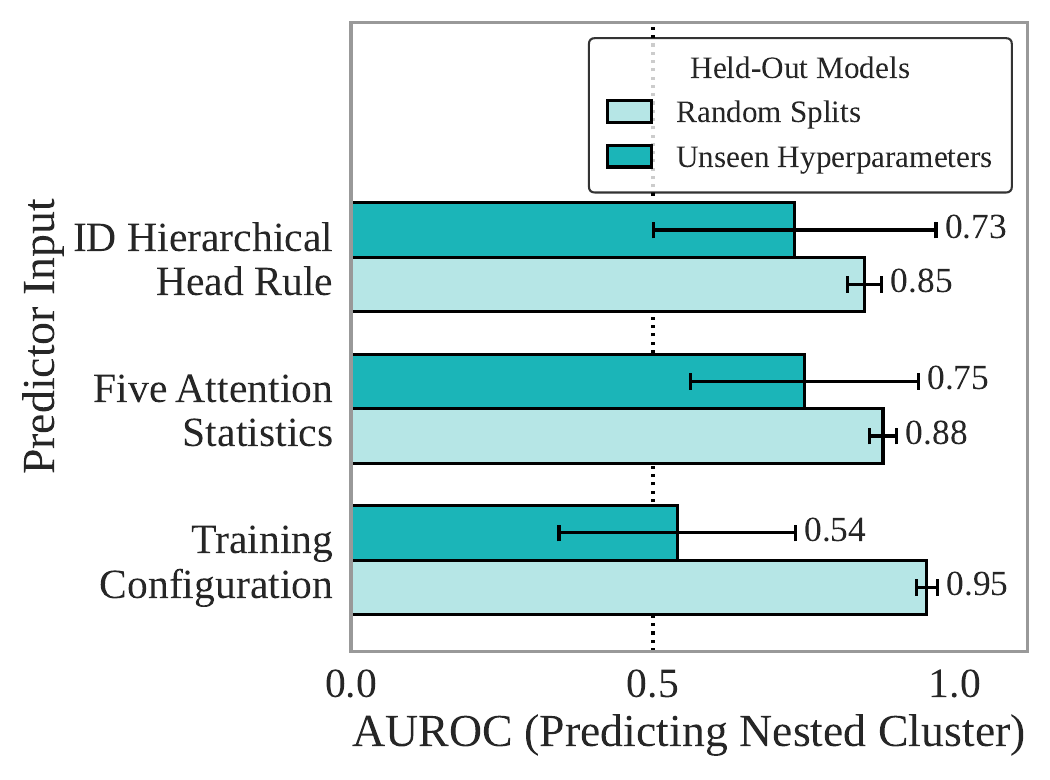}
\end{center}
\caption{\textbf{ID internals predict which unseen models follow \balanced{}; hyperparameters do not if their setting is unseen.} Each bar is the AUROC for predicting that a held-out Dyck model lies in the \balanced{} region of Fig.~\ref{fig:tsne-dyck} from ID information only: the ID hierarchical-head rule, five attention statistics, or the hyperparameters alone. Light bars: fit on 150 random models, tested on the other 120 (50 splits). Dark bars: all models with one (depth, weight decay) setting held out (six scorable settings). Whiskers: $\pm 1$ SD across splits; dotted: chance.}\label{fig:heldout}
\end{figure}

\textbf{Predictors.} We use three predictors: (1) The ID hierarchical-head rule in Section~\ref{sec:predicting-ood}. We associated \balanced{} behavior with heads that track violations on at least 80\% of relevant ID sequences. (2) We also train a random forest based on \EOS{}-attention statistics on 1K ID sequences (not relying on considering Equation~\ref{eq:nested}---in particular, center of mass, mass on the last tenth of the sequence, largest weight, weight on \EOS{}, and mean self-attention over all positions summarized by  max, mean and min over heads---15 numbers overall). (3) We further compare to a baseline of hyperparameters alone: a logistic regression on depth, width and weight decay.

\textbf{Evaluation.} We evaluate on random splits: 150 training models and 120 held-out models, 50 draws. Unseen hyperparameters are 30 models sharing one (depth, weight decay) setting fit on others; six of the nine settings contain both \balanced{} and non-\balanced{} models and can be scored. Figure~\ref{fig:heldout} reports the mean and standard deviation over splits.

\textbf{Results.} In random splits both the ID hierarchical-head rule and the five attention statistics-based predictors reach AUROC 0.85 and 0.88, respectively. The hyperparameters alone do  better ($0.95$), as weight decay and depth strongly shape rule choice (Appx.~\ref{sec:factors}). If we hold out a hyperparameter setting, hyperparameters fall to chance (0.54) while the ID head rule and attention statistics retain predictive power at $0.73$ and $0.75$ AUROC, respectively. The same two predictors transfer to the question-formation population ($n=79$; target: the \textsc{Hierarchical} region of Fig.~\ref{fig:tsne-qf}; all models share one configuration, so only random splits apply; 5-fold cross-validation): the Main Auxiliary-Detecting head reaches AUROC 0.76 and the same five statistics, computed at the position preceding the question, 0.72. Hierarchical heads are a readable and interpretable instance of ID internals as a predictive signal of OOD generalization.

\section{Breakdown of Hierarchical Heads by Layer and Weight Decay}\label{apx:breakdown}

\begin{figure*}[ht]
    \centering
    \begin{subfigure}{0.4\textwidth}
    \includegraphics[width=\linewidth]{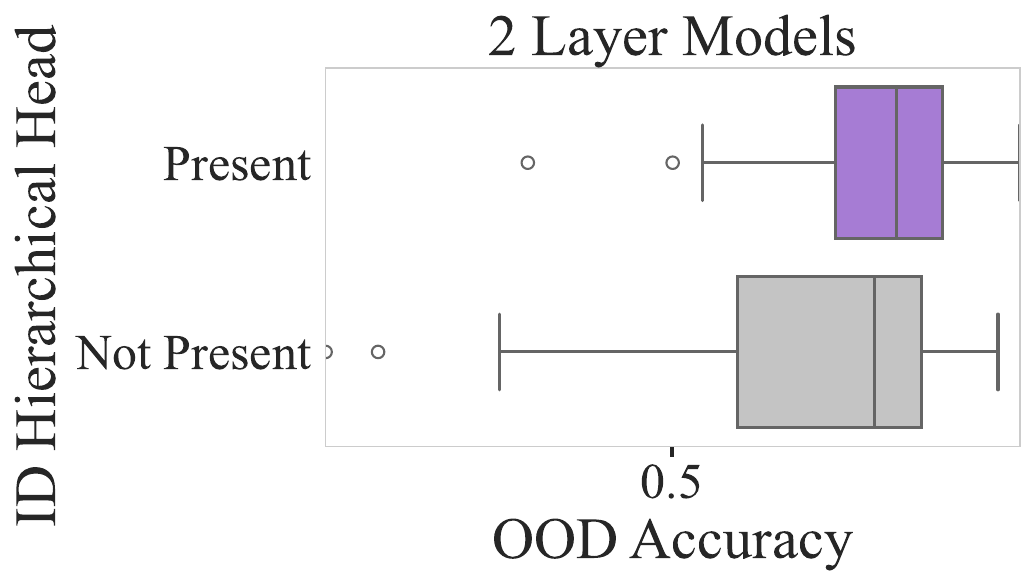}
    \end{subfigure}
    \hfill
    \begin{subfigure}{0.4\textwidth}
    \includegraphics[width=\linewidth]{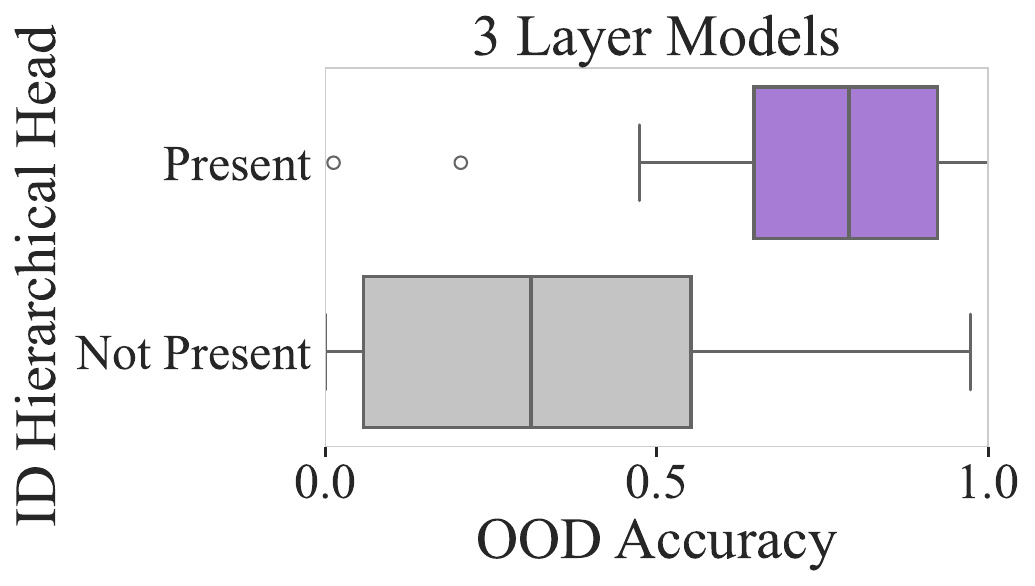}
    \end{subfigure}
    \caption{Last OOD test accuracy of 2- and 3-layer models with and without ID hierarchical heads.}
    \label{fig:boxplot_depth_heads_2layer}
\end{figure*}

Models with ID hierarchical heads consistently have higher OOD accuracy across both 2- and 3-layer models, indicating that model internals can provide additional predictive power to hyperparameters alone (Figure \ref{fig:boxplot_depth_heads_2layer}). Particularly for 3-layer models, which have the greatest diversity in OOD performance, the presence of ID hierarchical heads in a model provides additional insight into predicting hierarchical \balanced{}-like generalization.

Model internals continue to improve predictive power even when fixing both depth and weight decay simultaneously. Some hyperparameter combinations lead to one rule or the other relatively consistently: for example, ``1-layer, weight decay $>$ 0'' is completely predictive of \matched{} (see Figure \ref{fig:depth}, Appendix \ref{sec:model_depth}). However, in hyperparameter settings with diverse OOD behavior, presence of hierarchical heads is predictive of \balanced{} generalization behavior.

\begin{figure*}[!ht]
\begin{center}
    \includegraphics[width=0.6\linewidth]{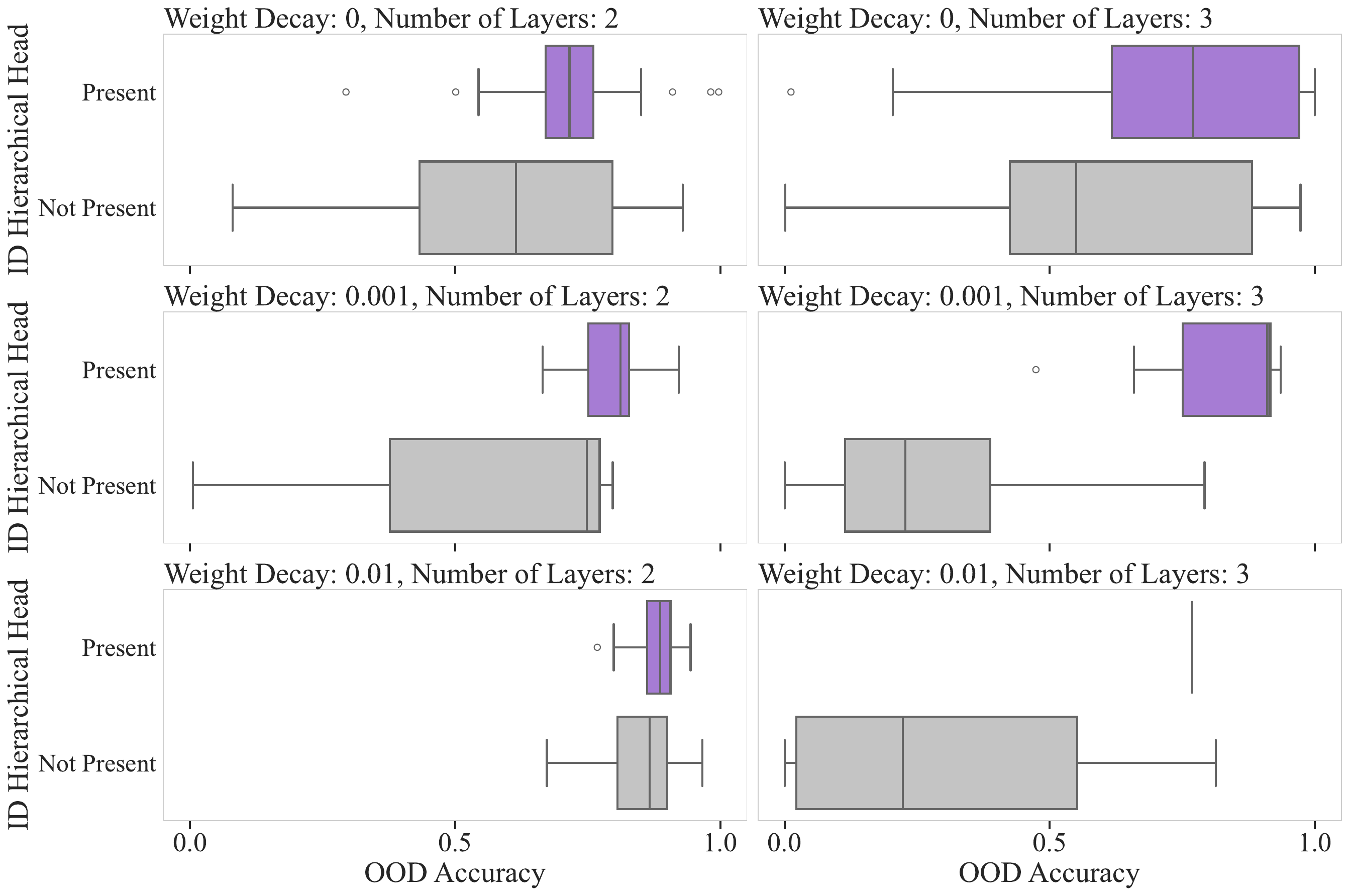}
\end{center}
\caption{OOD accuracy of 2- and 3-layer models with and without ID hierarchical heads, by number of layers and weight decay value. Over the four populations, (1) 2- and 3-layer models with non-zero weight decay, (2) all 2-layer models, (3) all 3-layer models, and (4) 3-layer models with 0.001 WD, a Mann--Whitney U test shows a significant difference in OOD accuracy distributions.}\label{fig:breakdown}
\end{figure*}

 In Figure \ref{fig:breakdown}, consider any multilayer setting producing a diverse range of OOD accuracy values, including values both above and below 50\% (in other words, any setting except the ``2-layer, 0.01 weight decay'' setting, where all models learn \balanced{}). In every such setting, models without hierarchical heads have a median OOD accuracy that falls at or below the bottom quartile of the score for models with such heads. For some settings, the distributions barely overlap at all. 

Thus, even if we consider the effect of hyperparameter settings,  the presence of a hierarchical (i.e., depth-tracking) head is highly informative. Of course, some of these settings might lead to more hierarchical runs \emph{because} they enable the learning of hierarchical mechanisms, meaning that the effect of hierarchical heads is far greater than any one setting would suggest.

\section{Effects of Causal Intervention on Attention}
\label{apx:corrs}
Our causal experiments involve uniform ablation of the attention distribution in all attention heads (Section \ref{sec:ablation}). We ablate all attention in order to uniformly and symmetrically intervene on all models. This is in contrast to ablating exclusively hierarchical heads, which would require us to compare models on an ``unequal footing,'' in the sense that some models would have 0 heads ablated, some models 1 head ablated, and some models 2 or more. Ablating all attention allows us to definitively eliminate all influence from hierarchical heads without introducing asymmetric interventions.

\begin{figure*}[!ht]
\begin{center}
        \includegraphics[width=0.65\linewidth]{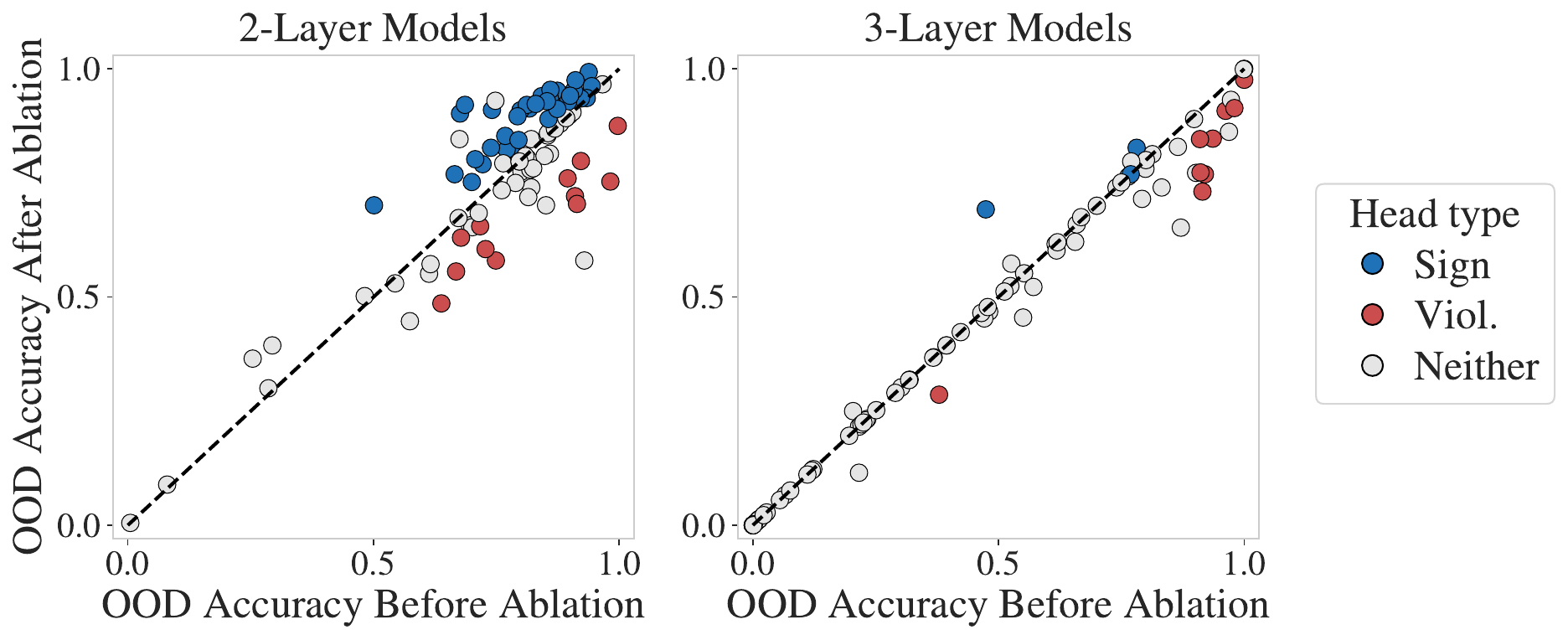}
\end{center}
\caption{OOD accuracy before and after applying uniform attention ablation to one head in each model. For each model, we plot the head whose ablation affects OOD accuracy most (in absolute value), colored by whether it is a sign-matching or violation-detecting head (or neither). In comparison with full attention ablation (Figure \ref{fig:scatter_plot_multilayers}), effects on OOD accuracy are smaller. However, our analysis of differing effects of ablating different types of hierarchical heads remains the same as in Section \ref{sec:ablation-results}.}\label{fig:single-heads}
\end{figure*}

 We also examined the effects of ablating individual heads one at a time. We found effects that were generally very similar (if weaker), in comparison with full attention ablation; in particular, in models with 2 or more hierarchical heads, full attention ablation tended to affect OOD accuracy more strongly than one-at-a-time attention ablation. Ultimately, one-head-at-a-time ablation demonstrates the  same trend as full attention ablation: ablating violation-detecting heads reduces \balanced{} behavior, while ablating sign-matching heads increases this behavior. See Figure \ref{fig:single-heads}. Mean ablation---replacing every head's attention pattern with its dataset-mean pattern over the OOD test set, either all heads at once or one head at a time---gives the same result as its uniform counterpart. In particular, across the 180 2- and 3-layer models, models with OOD sign-matching heads become more hierarchical when those heads are ablated, while models with violation-detecting heads become more linear.

\section{Change in OOD and ID Accuracy After Ablating Heads}\label{apx:head-ablation}

\begin{figure*}[ht]
    \centering
    \begin{subfigure}{0.49\textwidth}
    \includegraphics[width=0.7\linewidth]{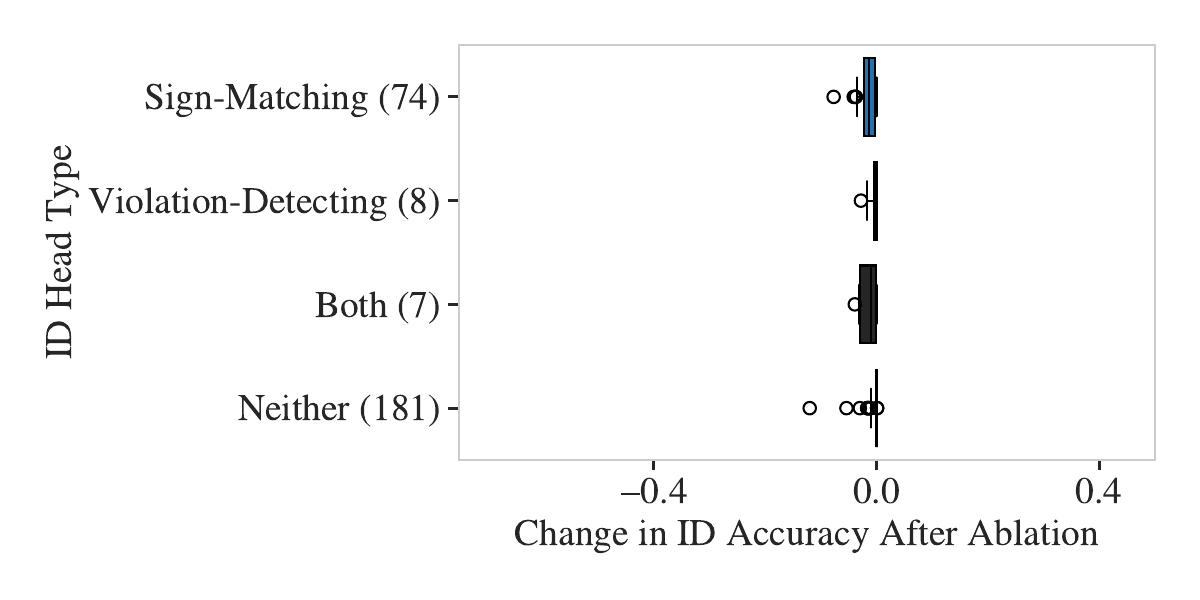}
    
    \end{subfigure}
    \hfill
    \begin{subfigure}{0.49\textwidth}
    \includegraphics[width=0.7\linewidth]{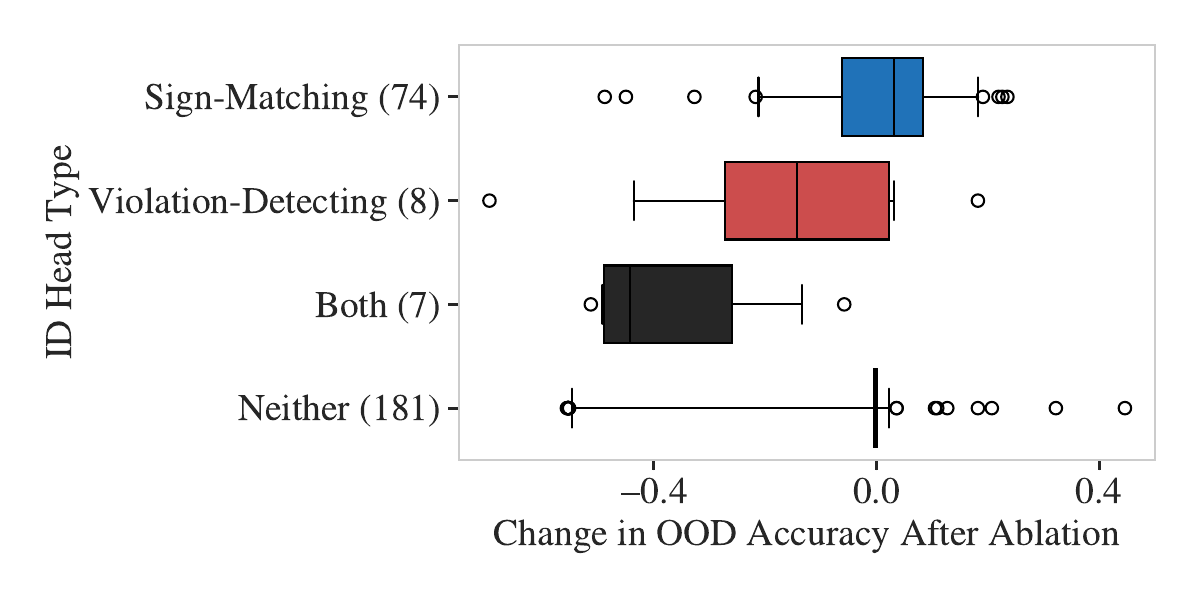}
    \end{subfigure}
    \caption{For heads classified by type based on their ID behavior, accuracy ID and OOD after ablation subtracted by baseline. Ablation has little impact on ID accuracy, and tends to decrease OOD accuracy across ID sign-matching and violation-detecting heads. The number of models in each category is included in parentheses after the label.}
    \label{fig:id-head-ablate}
\end{figure*}

\begin{figure*}[ht]
    \centering
    \begin{subfigure}{0.49\textwidth}
    \includegraphics[width=0.7\linewidth]{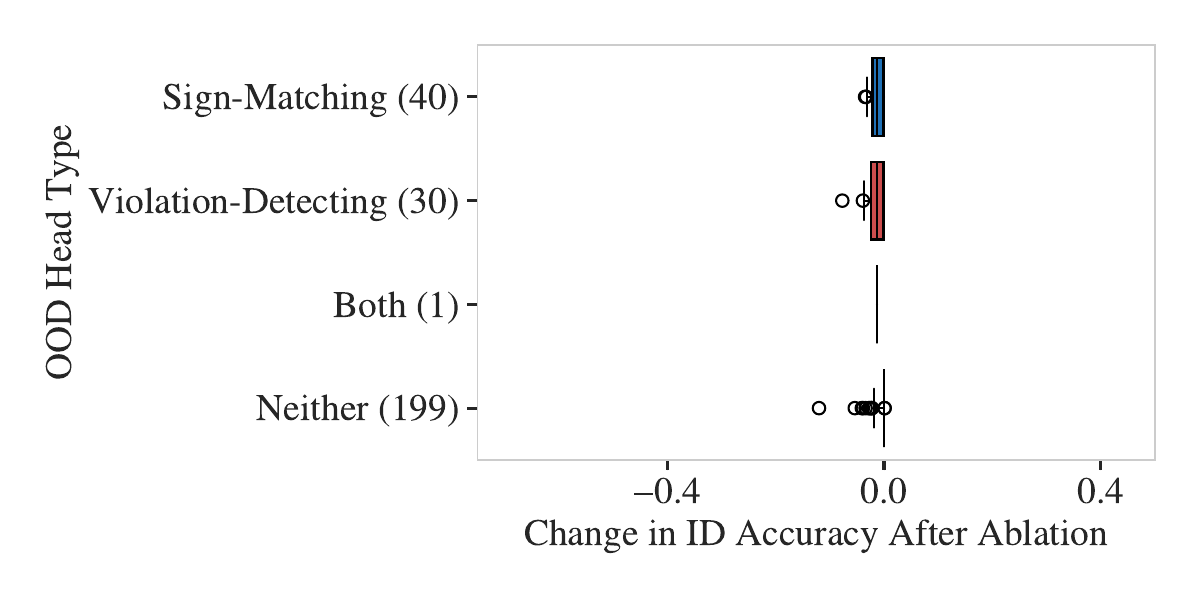}
    \end{subfigure}
    \hfill
    \begin{subfigure}{0.49\textwidth}
    \includegraphics[width=0.7\linewidth]{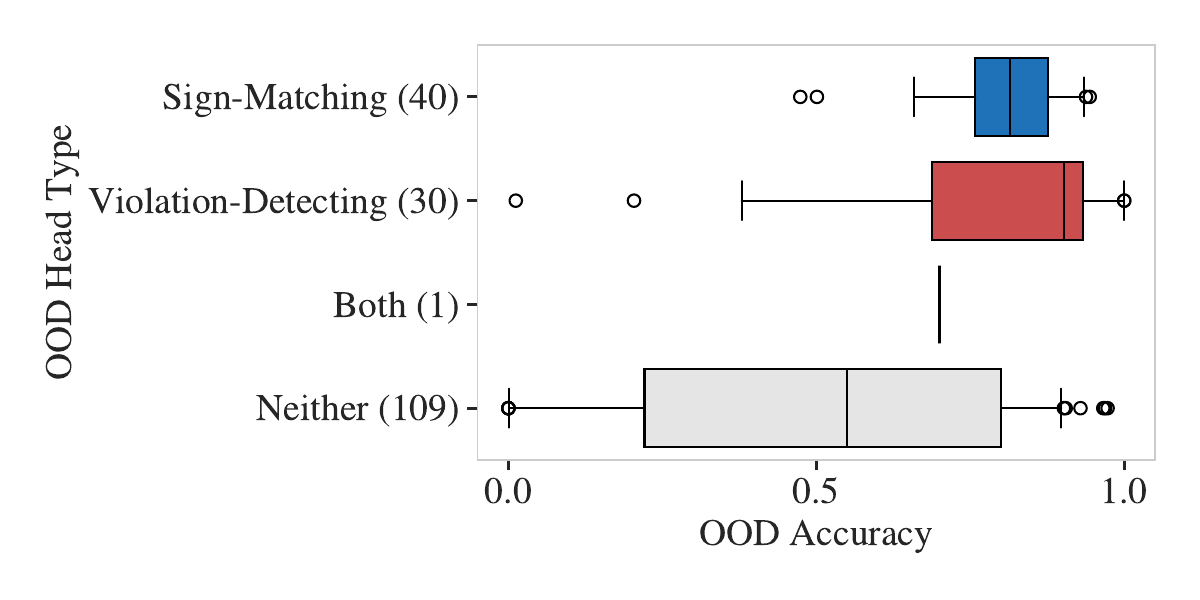}
    \end{subfigure}
    \caption{For heads classified by type based on their OOD behavior, accuracy ID and OOD after ablation subtracted by baseline. Ablation has little impact on ID accuracy, but ablating violation-detecting OOD heads decreases and ablating sign-matching OOD heads increases OOD accuracy. The number of models in each category is included in parentheses after the label.}
    \label{fig:ood-head-ablate}
\end{figure*}

Ablating hierarchical heads has little to no effect on ID accuracy (Figures \ref{fig:id-head-ablate} for ID hierarchical heads and \ref{fig:ood-head-ablate} for OOD hierarchical heads, left panels). The maximum effect of ablating an ID or OOD head type on the ID data is 77/1000, both for violation-detecting heads, but the median impact is around 10/1000 for all classified head types. Across ID head types, ablation tends to decrease OOD accuracy, but for OOD head types, as seen in Figure \ref{fig:scatter_plot_multilayers}, ablating violation-detecting heads decreases while ablating sign-matching heads increases OOD accuracy (Figures \ref{fig:id-head-ablate} and \ref{fig:ood-head-ablate}, right panels).

\section{Question Formation Setting}\label{apx:qf}

\subsection{Task and Data}
Our autoregressive setting is question formation in English~\cite{mccoy-etal-2020-syntax,qin2024itreedatadrives}: an autoregressive model reads a declarative sentence followed by the special token marker {\color{darkblue}{\texttt{quest}}} and produces the corresponding yes/no question. The correct transformation requires identifying the \emph{matrix} (root) auxiliary and moving it to the front:
\begin{quote}
\textit{My dog does like the ravens that do dance.}~{\color{darkblue}\texttt{quest}} \textit{Does my dog like the ravens that do dance?}
\end{quote}
Two rules yield the same output on training and ID validation:
\begin{itemize}[leftmargin=*, topsep=0pt, parsep=0pt, itemsep=2pt, partopsep=0pt]
\item \textbf{Linear rule:} move the \emph{first} auxiliary in the sentence to the front.
\item \textbf{Hierarchical rule:} move the \emph{matrix} (root-clause) auxiliary to the front.
\end{itemize}
ID examples are constructed so that the first auxiliary \emph{is} the matrix auxiliary, so both rules predict the same output token; OOD examples place an auxiliary inside a relative clause that precedes the matrix auxiliary, making the two rules diverge. Table~\ref{tab:strategies} contrasts the two settings of this paper.
\begin{table*}[t]
\centering
\scriptsize
\setlength{\tabcolsep}{6pt}
\renewcommand{\arraystretch}{1.25}
\begin{tabular}{@{}c l l l@{}}
\toprule
 & & \textbf{Dyck-1} \emph{(classification)}
   & \textbf{Question formation} \emph{(autoregressive)} \\
\midrule
\multirow{2}{*}{\textbf{ID}}
  & Input
    & \makecell[l]{\open\close\open\open\close\close \\ \open\open\open\close\open}
    & My dog \auxM{does} like the ravens that \auxE{do} dance. \\
\addlinespace[2pt]
  & Output
    & \makecell[l]{\cmark\,\textsc{true} \\ \xmark\,\textsc{false}}
    & \auxM{Does} my dog like the ravens that \auxE{do} dance? \\
\midrule
\multirow{3}{*}{\textbf{OOD}}
  & Input
    & \open\close\close\open\open\close
    & My dog who \auxE{doesn't} sing \auxM{does} dance. \\
\addlinespace[2pt]
  & \textsc{Equal-Count} / linear
    & \cmark\,\textsc{true}
    & \auxE{Doesn't} my dog who sing does dance? \\
\addlinespace[2pt]
  & \textsc{Nested} / hierarchical
    & \xmark\,\textsc{false}
    & \auxM{Does} my dog who doesn't sing dance? \\
\bottomrule
\end{tabular}
\caption{Two diagnostics for \emph{which} rule a model has actually learned.
ID examples are consistent with both candidate rules; OOD examples force them
apart. In Dyck-1, the linear (\matched{}) and hierarchical (\balanced{}) rules
disagree about whether a non-nested-but-equal-count sequence is \texttt{True}. In
question formation, they disagree about \emph{which} auxiliary moves: in this example, the
embedded \auxE{doesn't} (linear) or the matrix \auxM{does} (hierarchical).}
\label{tab:strategies}
\end{table*}

OOD accuracy measures hierarchical generalization: we report it as $P(\text{hier})$, the mean probability the model assigns to the matrix (hierarchical) auxiliary at the fronting position. 100\% means the model consistently moves the matrix auxiliary; 0\% means it consistently moves the first auxiliary.  
\subsection{Models and Population}
We analyze 79 distinct autoregressive Transformer language models in this population, following the training procedure of \citet{qin2024itreedatadrives}. Each is a $6$-layer decoder-only (causal) Transformer with $8$ attention heads, model dimension $512$, feed-forward dimension $2048$, dropout $0.1$, and tied input/output embeddings (vocabulary size $71$, $\approx\!19$M parameters). The model reads the declarative followed by the {\color{darkblue}\texttt{quest}} marker with causal self-attention and predicts the fronted auxiliary as the next token. All models are trained with Adam: learning rate $10^{-4}$, $10{,}000$ warm-up steps, a two-phase learning-rate schedule, no weight decay or label smoothing, gradient-norm clipping at $1.0$, and batch size $8$. All are evaluated at the $300$K-step checkpoint and reach perfect ID validation accuracy. Because the models share architecture and objective and differ essentially only in random seed (weight initialization and data ordering), we can compare their OOD behavior directly. The heads we analyze are causal self-attention heads; ``first three layers'' refers to the first three of the six layers.

\subsection{Main Auxiliary-Detecting ID Head}\label{apx:qf-heads}

\paragraph{Pattern.} A \textbf{Main Auxiliary-Detecting head} is an attention head in the \emph{first three model layers} whose attention from the {\color{darkblue}\texttt{quest}} token---the final input token, at which the model begins generating the question---lands on the \emph{matrix} (root-clause) auxiliary, i.e.\ the auxiliary that the hierarchical rule fronts. For an ID sentence ``My unicorn does move the dogs that do wait.'' the matrix auxiliary is \textit{does} (not the embedded \textit{do}); for the OOD sentence ``My unicorn who doesn't sing does move.'' it is still \textit{does} (not the embedded \textit{doesn't}).

\paragraph{Detection details.} For each (model, layer, head) with the layer among the first three, we mark an ID validation example as \emph{detected} if the attention weight $a({\color{darkblue}\texttt{quest}} \to \text{matrix\_aux}) > 0.2$; a head qualifies as Main Auxiliary-Detecting if it is detected on at least $80\%$ of the ID validation set. A model carries the head type if at least one head in its first three layers qualifies. The two thresholds are robust: sweeping the per-token attention cutoff (from $0.025$ to $0.40$) and the persistence threshold (from $0.5$ to $1.0$), the gap in mean OOD hierarchical accuracy between models that carry the head and models that do not stays positive and generally statistically significant. At the reported cutoff in Figure \ref{fig:headtype_predicts_ood} ($>0.2$ on $\geq 80\%$ of ID examples), $59$ of $79$ models ($75\%$) carry at least one Main Auxiliary-Detecting head (a one-sided Mann--Whitney U test rejects equality at $p = 6.9\times10^{-3}$).

\subsection{Ablation Does Not Surface a Dyck-Style Decoupling}\label{apx:qf-ablation}

\begin{table}[!ht]
\centering
\footnotesize
\setlength{\tabcolsep}{3pt}
\renewcommand{\arraystretch}{1.15}
\resizebox{\linewidth}{!}{\begin{tabular}{@{}lrrrrr@{}}
\toprule
Ablation & mean $\Delta$OOD & median $\Delta$OOD & 95\% CI & $\Delta$ID & $p_{\text{Wilc.}}$ \\
\midrule
Mean            & $+0.002$ & $+0.000$ & $[-0.004, +0.009]$ & $0.000$ & $0.74$ \\
Uniform attn.   & $-0.007$ & $-0.000$ & $[-0.020, +0.004]$ & $0.000$ & $0.23$ \\
\bottomrule
\end{tabular}}
\caption{Change in OOD hierarchical accuracy when ablating Main Auxiliary-Detecting heads (all target heads together), for the $n=59$ models with at least one. Two ablation types (mean, uniform-attention) and one-sided Wilcoxon test. Both effects are within a percentage point of zero, neither reaches $p<0.05$, and ID validation accuracy is unchanged. One-at-a-time ablation gives the same result.}
\label{tab:qf-ablation-types}
\end{table}

We ablate Main Auxiliary-Detecting heads in every model that carries at least one ($n=59$) and measure the change in OOD hierarchical accuracy. To check the result is not an artifact of a specific ablation type, we apply the same ablations we use in the Dyck setting. We consider two ablation types per model, each applied both to all target heads together and to each target head one at a time:
\begin{itemize}[leftmargin=*, topsep=0pt, parsep=0pt, itemsep=2pt, partopsep=0pt]
\item \textbf{Mean ablation:} replace the head's attention pattern with its dataset-mean pattern, computed on the evaluation set.
\item \textbf{Uniform-attention ablation:} replace the head's attention weights with uniform attention over allowed (causally unmasked) positions.
\end{itemize}
Table \ref{tab:qf-ablation-types} reports the per-type distribution of $\Delta$ (OOD hier. acc.) across the 59 ablated models. Both ablation types yield a near-zero mean effect, whether the target heads are ablated together or one at a time. Crucially, ID validation accuracy is unchanged under every intervention---all 59 models retain perfect ID accuracy---so the heads are not load-bearing for the in-distribution task either. The qualitative finding is robust: the Main Auxiliary-Detecting head correlates with hierarchical OOD behavior at ID but does not load-bear under these interventions. The surprising correlational/causal \emph{decoupling} we report in Dyck-1 (where sign-matching attention predicts hierarchy yet ablating it \emph{increases} hierarchy) is not a general result but a proof of existence.

\section{AI Statement}

AI assistance was used for some analysis, wording, and visualizations. All AI generated code was manually verified. 

\clearpage

\end{document}